\documentclass[11pt]{article}
	
	\newcommand{\blind}{0}
	
 \makeatletter
    \renewcommand\section{\@startsection{section}{1}{\z@}%
                                       {-3.5ex \@plus -1ex \@minus -.2ex}%
                                       {2.3ex \@plus.2ex}%
                                       {\normalfont\normalsize\fontfamily{phv}\fontsize{16}{19}\bfseries}}
    \renewcommand\subsection{\@startsection{subsection}{2}{\z@}%
                                         {-3.25ex\@plus -1ex \@minus -.2ex}%
                                         {1.5ex \@plus .2ex}%
                                         {\normalfont\fontfamily{phv}\fontsize{14}{17}\bfseries}}
    \renewcommand\subsubsection{\@startsection{subsubsection}{3}{\z@}%
                                        {-3.25ex\@plus -1ex \@minus -.2ex}%
                                         {1.5ex \@plus .2ex}%
                                         {\normalfont\normalsize\fontfamily{phv}\fontsize{12}{17}\selectfont}}
    \makeatother
	
	\usepackage{amsmath}
	\usepackage{graphicx}
	\usepackage{enumerate}
	\usepackage{xcolor}
    \usepackage{amssymb}  

    \usepackage{threeparttable} 
    \usepackage{indentfirst}
	\usepackage{natbib} 
	\usepackage{url} 
    \usepackage{algorithm}
    \usepackage[noend]{algpseudocode}
    \usepackage[super]{nth}
    \usepackage{hyperref}
    \usepackage{flafter}   
    \usepackage{placeins}
    \usepackage{needspace}   
    \usepackage{siunitx}
	\usepackage{todonotes}

	\usepackage{lipsum} 
        \usepackage{tablefootnote}
\begin{document}

		\def\spacingset#1{\renewcommand{\baselinestretch}%
			{#1}\small\normalsize} \spacingset{1}
        \if0\blind
		{
			\title{\bf Discovering Heuristics with Large Language Models (LLMs) for Mixed-Integer Programs: Single-Machine Scheduling
   }
			\author{İbrahim Oğuz Çetinkaya$^1$, {\.I.} Esra B{\"u}y{\"u}ktahtak{\i}n$^1$\thanks{Corresponding author: esratoy@vt.edu} \\
            Parshin Shojaee$^2$, Chandan K. Reddy$^2$ \\
            $^1$\small Grado Department of Industrial and Systems Engineering, Virginia Tech, Blacksburg, VA \\$^2$ \small Department of Computer Science, Virginia Tech, Arlington, VA
\\}
			\date{\today}
			\maketitle
		} \fi

		\if1\blind
		{

            \title{\bf \emph{IISE Transactions} \LaTeX \ Template}
			\author{Author information is purposely removed for double-blind review}
			
\bigskip
			\bigskip
			\bigskip
			\begin{center}
				{\LARGE\bf \emph{IISE Transactions} \LaTeX \ Template}
			\end{center}
			\medskip
		} \fi
		\bigskip

	\spacingset{1.5} 
\vspace{-1cm}

\begin{abstract}
Our study contributes to the scheduling and combinatorial optimization literature with new heuristics discovered by leveraging the power of Large Language Models (LLMs). We focus on the single-machine total tardiness (SMTT) problem, which aims to minimize total tardiness by sequencing $n$ jobs on a single processor without preemption, given processing times and due dates. We develop and benchmark two novel LLM-discovered heuristics, the EDD Challenger (EDDC) and MDD Challenger (MDDC), inspired by the well-known Earliest Due Date (EDD) and Modified Due Date (MDD) rules. In contrast to prior studies that employed simpler rule-based heuristics, we evaluate our LLM-discovered algorithms using rigorous criteria, including optimality gaps and solution time derived from a mixed-integer programming (MIP) formulation of SMTT. We compare their performance against state-of-the-art heuristics and exact methods across various job sizes (20, 100, 200, and 500 jobs). For instances with more than 100 jobs, exact methods such as MIP and dynamic programming become computationally intractable. Up to 500 jobs, EDDC improves upon the classic EDD rule and another widely used algorithm in the literature. MDDC consistently outperforms traditional heuristics and remains competitive with exact approaches, particularly on larger and more complex instances. This study shows that human–LLM collaboration can produce scalable, high-performing heuristics for NP-hard constrained combinatorial optimization, even under limited resources when effectively configured.
\end{abstract}

\section{Introduction} 
\label{s:intro}

    Mixed Integer Programs (MIPs) represent a class of optimization problems that involve both integer and continuous decision variables. These problems are notoriously challenging because of their combinatorial nature, and many formulations in this domain are classified as NP-hard. Among these, the single-machine total tardiness (SMTT) problem, which seeks to minimize total tardiness, is a foundational problem at the intersection of sequencing and scheduling within the broader MIP framework. Its complexity and NP-hardness, proven by \cite{Du1990-nphard}, make it a significant testbed for both exact and heuristic solution methods. Following the standard three-field scheduling notation of \cite{graham1979} and \cite{pinedo2012}, 
the single-machine total tardiness (SMTT) problem is denoted as \(1 \| \sum T_j\). 
In this notation, the first field ``1'' specifies that there is a single machine, 
the second field ``\(\|\)'' indicates the absence of additional constraints such as preemption 
or machine environments, and the third field ``\(\sum T_j\)'' denotes the objective of minimizing 
the sum of job tardiness. A set of \(n\) jobs, indexed by \(j = 1, \ldots, n\), must be processed 
without preemption on a single machine. Each job \(j\) has a completion time \(C_j\) and a due date \(d_j\). 
The objective is to minimize the total tardiness \(\sum_{j=1}^n T_j\), 
where the tardiness of job \(j\) is defined as \(T_j = \max \{ 0, C_j - d_j \}\).

    The single-machine total tardiness (SMTT) problem constitutes a ``building block" to more complex scheduling problems (\cite{koulamas2023classification}). Given that ``meeting the job due dates" is one of the most common criteria in many situations, SMTT provides a tractable model to analyze various contexts and applications (\cite{baker2018}).  Moreover, minimizing the total tardiness objective ensures that none of the jobs wait in the queue for too long and helps construct more balanced schedules, which is a practical concern. For these reasons, SMTT has received a lot of interest in scholarly circles as a practical problem. It is central to the domain of production planning (\cite{bauer2000}) with various extensions (\cite{agentis2007}, \cite{wang2022minimizing}, and many more). For example, SMTT serves as a core problem in semiconductor manufacturing to respond to customer demand in a timely manner (\cite{gupta2005sc}), and a way to model envelope and videotape production (\cite{li1997}). In short, SMTT is a vital problem for both the practical applications in the industry and the theoretical advances in the scheduling domain, and hence it draws close attention from practitioners and academicians.

  Large Language Models (LLMs) have emerged as powerful tools capable of completing complex tasks (\cite{zhao2023survey, shanahan2024talking}), however, they are still far away from achieving algorithmic innovation by themselves. Moreover, when embedded in rigorous evaluation frameworks, LLMs are able to generate structured outputs and propose algorithmic ideas. While LLMs alone are not sufficient for reliable scientific discovery, they can be highly effective when paired with iterative validation and refinement. FunSearch (\cite{funsearch}) exemplifies this approach by coupling LLMs with an evolutionary pipeline based on genetic programming, where candidate programs are mutated and evaluated iteratively to improve performance. Our work builds on this paradigm and tailors it to the scheduling domain, showing that LLMs can be guided to discover new heuristics for NP-hard problems that outperform both classical heuristics and exact methods.

    A vast body of literature has evolved over decades for the single-machine minimize tardiness scheduling problem, focusing on both advanced methodologies to accelerate exact methods and approximate solution methods that deliver near-optimal solutions with computational efficiency. Our study contributes to the fields of combinatorial optimization and scheduling by leveraging the computational power and flexibility of Large Language Models (LLMs) to discover new construction heuristics.

    We implement an LLM-backed algorithmic discovery pipeline, translating the scheduling problem into computer code, and using LLMs to iteratively enhance algorithms in a sandbox environment. The translation process involves two key components: (1) an initial heuristic method provided to the LLM as a foundation, and (2) a systematic evaluator to ensure feasibility and assess total tardiness over a dataset. The LLM refines the initial heuristic, while the systematic evaluator verifies feasible outputs and measures performance.

    To generate new heuristics, we provide the LLM with three initial heuristics: Earliest Due Date (EDD), Shortest Processing Time (SPT), and more advanced Modified Due Date (MDD). The EDD-based pipeline surpasses one heuristic from the literature but falls short of matching the best-known state-of-the-art algorithms. The SPT-based pipeline does not produce a well-performing heuristic; however, the best function discovered through the MDD-based pipeline surpasses both the original heuristic and the best-performing heuristic used in this study, advancing the state-of-the-art. The proposed algorithms are further evaluated on larger-scale problem instances, where the LLM-discovered heuristic consistently outperforms its competitors, demonstrating its robustness and scalability.

    Our study demonstrates that LLMs can be effectively leveraged to discover heuristics for mathematical problems, highlighting the flexibility and computational power of these novel techniques when combined with domain expertise. Our methodology is particularly valuable, as it uses small training instances - starting with 25 job problems - to generate heuristics that perform exceptionally well at larger scales, including 500 job instances. Another distinguishing aspect of our work is the emphasis on the clarity and reproducibility of results. We present the developed algorithms using our LLM-backed framework as pseudocode -and as Python code in Github-, while certain terms in the proposed algorithms may diverge from conventional scheduling logic, the decision-making process remains transparent and interpretable. This stands in contrast to the black-box approaches commonly seen in the machine learning and operations research literature.

\section{Literature Review}
\label{s:lit_rev}
    In this section, we review the relevant literature related to both the single-machine total tardiness scheduling problem and scientific discovery using LLMs. In the SMTT problem, a set of \(n\) jobs must be processed on a single machine without preemption, with the objective of minimizing the total tardiness \(\sum T_j\), where \(T_j = \max\{0, C_j - d_j\}\). The NP-hardness of the problem remained an open question for a long time. Although \cite{lawlerDecomposition} proved that the weighted version of the problem is NP-hard much earlier, the NP-hardness of the original problem was not established until \cite{Du1990-nphard}'s proof. However, efforts to develop solution methods have persisted for much longer. One of the pioneering contributions in this field is by \cite{Held1962}, who propose the use of dynamic programming (DP) to tackle sequencing problems. They observe that as the problem size increases, the DP struggles to remain computationally feasible. Similarly, \cite{bakerDP} apply DP to sequencing problems with precedence relations and raise concerns about the curse of dimensionality, echoing the challenges noted by \cite{Held1962}.

    Two major theoretical advancements that have inspired many subsequent studies come from \cite{Emmons1969} and \cite{lawlerDecomposition}. \cite{Emmons1969} introduces the dominance criterion to compare jobs and determine their relative positions in an optimal schedule. Building on this, Lawler proposes a decomposition algorithm based on the rightmost assumption, suggesting that the job with the longest processing time should be scheduled as late as possible in the optimal sequence, allowing for a pseudo-polynomial solution approach. A significant extension comes from \cite{Abdurrezak1988}, who incorporate state-space modifiers into the DP formulation, narrowing the solution space and reducing computation times. Later, \cite{dellaCroce} refine Lawler’s decomposition conditions, while \cite{changDecomposition} propose a new decomposition rule that further advances the field.

    \cite{Potts1982} develop an exact method capable of solving problems with up to 100 jobs by leveraging the ideas of \cite{Emmons1969} and \cite{lawlerDecomposition}. They also contribute to the field by formalizing a data-generation schema to create problem instances. More recently, \cite{decompositionDL} employ LSTM networks to decompose the problem across varying sizes, from 5 to 800 jobs, demonstrating that their approximation approach achieves lower optimality gaps than state-of-the-art decomposition methods.

    In addition to dynamic programming, several studies model the problem as a Mixed Integer Programming (MIP) problem. \cite{Dyer1990} formulate the problem as an MIP, while \cite{Keha2009} present four MIP formulations and benchmark their computational performance. Their results show that the commonly used sequencing decision-based model is outperformed by a model based on the principle of positional assignment, which is further strengthened by a set of valid inequalities (discussed in Section 3). Another popular method for solving the problem is branch-and-bound. \cite{Kianfar2012} apply branch-and-bound to minimize the quadratic earliness/tardiness objective, using Lagrangian relaxation to derive lower bounds and a two-step heuristic to compute upper bounds.

    The recent work of \cite{shang2021} introduces the branch-and-memorize algorithm, an enhancement of traditional branch-and-bound methods that leverages memorization to reduce redundant computations and accelerate pruning. Although this approach increases memory usage, it significantly reduces solution times. The algorithm employs three memorization schemes: (1) solution memorization reuses optimal sub-problem solutions, (2) passive node memorization prunes dominated branches, and (3) predictive node memorization integrates local searches to manage partial solutions. Applied to various scheduling problems, including the single-machine total tardiness problem examined in this paper, the branch-and-memorize algorithm demonstrates the capability to solve instances with up to 1200 jobs, far exceeding the 200-job limit of conventional methods. This innovative framework highlights the effectiveness of combining memorization techniques with search strategies to address larger and more complex problems efficiently.

    \indent Apart from exact methods, valuable heuristic algorithms have been developed, particularly for real-world applications where near-optimal solutions are needed within shorter solution times. \cite{mdd} propose the MDD rule, which sequences jobs in non-increasing order of \(\max\{d_j, t + p_j\}\), where \(d_j\) are the due dates, \(p_j\) are the processing times, and \(t\) is the current time. \cite{Bean} further enhance the MDD rule by introducing an augmentation tactic to improve its performance. \cite{PSK} present a robust heuristic algorithm built on a series of logical rules to guide scheduling decisions. Another effective heuristic of \cite{Panneerselvam2006} focuses on sorting jobs according to their slack times. For a comprehensive review of the single-machine total tardiness scheduling problem, we refer to \cite{koulamas2010review}'s seminal work.

    \indent The study titled ``Mathematical discoveries from program search with large language models'' by \cite{funsearch} represents a significant advancement in the field of mathematical equation discovery. The proposed FunSearch framework integrates LLMs with evolutionary algorithms to generate new equations. The framework consists of a pre-trained LLM and a systematic evaluator, both provided in the form of computer code. The LLM is initialized with an initial solution function, which iteratively evolves, while the systematic evaluator ensures the feasibility and evaluates the performance of the generated solutions.

    The pipeline begins by modifying the initial solution function and testing it on a dataset. Feasible programs are stored in a program database, forming the basis for future iterations. Using a ``best-shot prompting technique '', new solution functions are derived from the database and fed back to the LLM for further refinement. The evolutionary process follows an island-based approach, allowing the solution to improve over time. This methodology enabled the authors to discover a new heuristic for the online bin-packing problem and novel cap-sets, leading to results previously not achieved in the literature.

    \indent Another novel approach is introduced by \cite{evolutionofheur}, who propose a framework to generate heuristics using LLM within an evolutionary algorithm structure. Their method begins by evolving high-level heuristic ideas in natural language, which are then translated into computer code. Two prompting strategies are utilized: one focuses on exploring new heuristics, while the other refines existing ones. These strategies operate in parallel, iteratively improving the heuristic population over time.

    The framework demonstrates impressive results, outperforming state-of-the-art heuristics across several well-known combinatorial problems, including online bin-packing, flow-shop scheduling, and the traveling salesperson problem. In particular, the EoH heuristic surpasses the bin-packing solution from \cite{funsearch}, achieving better performance with less computational power.

    \indent Building on the funSearch concept, \cite{funBO} explore new acquisition functions (AFs) for Bayesian Optimization. These AFs, discovered through a similar pipeline, outperform standard AFs and show competitive performance against domain-specific alternatives. In a different field, \cite{molecularLLM} integrate LLMs with prior scientific knowledge to create a discovery pipeline called ``LLM4SD." Their framework, applied to 58 molecular prediction tasks, uncovers novel rules across multiple disciplines, including physiology, biophysics, physical chemistry, and quantum mechanics—insights previously unidentified by researchers.

    \cite{LLM-A*} leverage LLMs to enhance the well-known A* algorithm used in path planning for robotics. The authors utilize the reasoning capabilities of LLMs to reduce the computational burden and memory demands of A*, while retaining the core structure of the original algorithm to ensure the validity of the solution. The resulting hybrid algorithm, LLM-A*, outperforms traditional A* by improving efficiency without compromising accuracy.

    In recent advancements in the application of ML to solving MIP problems, several other notable studies have emerged. \citet{yilmaz2022learning} and \citet{yilmaz2022an} utilize the Long Short-Term Memory (LSTM), a type of recurrent neural networks (RNN), and encoder-decoder deep learning approaches to predict binary variables in capacitated lot-sizing and optimize sequential decision-making problems, respectively. \citet{yilmaz2022_multistageSP} extend the encoder-decoder prediction framework to learn solutions for multi-stage stochastic MIPs. \citet{pan2022deepopf} propose a Feasibility-Optimized Deep Neural Network for AC optimal power flow challenges. 
    
    \citet{bello2016neural} integrate neural networks with reinforcement learning (RL) for the Traveling Salesman Problem (TSP), optimizing RNN parameters via a policy gradient method that utilizes negative tour length as a reward. Finally, \citet{yilmaz2022_RL} develop a deep reinforcement learning (DRL) framework to tackle scenario-based stochastic programs \citet{buyuktahtakin2022stage, buyuktahtakin2023scenario} in two stages with a multiagent structure, while \citet{Bushaj2024} tackle multidimensional knapsack problems (MKP) via a novel DRL framework enhanced by a K-means algorithm and \citet{Bushaj2023} integrates RL with epidemic models for pandemic prevention decision-making.

    Our study offers several key contributions to the combinatorial optimization, scheduling and knowledge discovery literature.
    First, we introduce and benchmark two novel LLM-discovered heuristics, EDD Challenger (EDDC) and MDD Challenger (MDDC), developed based on the EDD and MDD rules. Second, our computational study rigorously evaluates the performance of LLM-discovered heuristics using MIP evaluation criteria, such as optimality gaps, and conducting a comprehensive comparative analysis against both state-of-the-art heuristics and exact methods. A key strength of our work is its rigorous experimental analysis, which covers a wide variety of single-machine scheduling instances generated by in and out-of-sample distributions, as well as benchmark instances from the literature across multiple scales (e.g., 20, 100, 200, and 500-job problems).  This extensive evaluation not only highlights the scalability of the LLM-discovered heuristics but also ensures their robustness across varying problem complexities. Unlike previous studies, such as \cite{funsearch}, which used only basic heuristics and smaller datasets, we compared MDDC and EDDC with more advanced heuristics and exact MIP and DP approaches, demonstrating that LLM-generated algorithms perform well in both small and large instances. Then, another important aspect of our study is the use of `prompt engineering' which is a novel technique frequently used in the industry to guide LLMs. Third, the two newly discovered heuristics, EDDC and MDDC, achieve smaller optimality gaps compared to well-known heuristic methods, especially for complex single-machine scheduling problems. The EDDC algorithm enhances EDD's performance, surpassing also Pannerselvam's performance. The MDDC, particularly the Augmented MDDC, consistently outperforms EDDC and traditional heuristics, offering competitive performance with exact methods, highlighting its practical utility for large-scale optimization problems. 
    \section{Model Formulation}
    In this section, we present mixed-integer programming (MIP) and dynamic programming (DP) models to solve the single-machine scheduling problem with the objective of minimizing total tardiness. Several alternative mathematical formulations exist for this problem, including time-indexed models \citep{abdul1990survey}, linear ordering formulations, and assignment-based formulations with positional date variables \citep{khowala2005comparison}. While each has its merits, we adopt a positional assignment-based MIP model introduced by \citep{Keha2009}, which is compact and effective for small- to medium-sized instances. This formulation lends itself well to benchmarking against learning-based heuristics and facilitates interpretability through assignment variables.

    \subsection{Mixed Integer Programming (MIP) Formulation}
    This section presents the notation and the mixed-integer programming (MIP) formulation to obtain optimal solutions. The approach implemented here, as described in \cite{Keha2009}, uses the principle of positional assignment, contrasting with the most commonly preferred sequencing decision methods in the literature.

    \subsubsection*{Nomenclature}
    
    \textbf{Indices and Sets:}
    \begin{itemize}
        \item[] $j \in N$: Jobs, where $N = \{1, 2, \ldots, n\}$
    \end{itemize}
    
    \textbf{Parameters:}
    \begin{itemize}
        \item[] $d_j$: Due date of job $j$
        \item[] $p_j$: Processing time of job $j$
        \item[] $M$: A large constant (sum of processing times)
    \end{itemize}
    
    \textbf{Decision Variables:}
    \begin{itemize}
        \item[] $u_{jk} \in \{0,1\}$: 1 if job $j$ is assigned to position $k$, 0 otherwise
        \item[] $c_k$: Completion time of the job at position $k$
        \item[] $C_j$: Completion time of job $j$
        \item[] $T_j$: Tardiness of job $j$
    \end{itemize}

        \subsubsection*{MIP Formulation}

        \[
        \text{Minimize} \quad \sum_{j \in N} T_j \label{eq:eq_0} 
        \]
        \begin{align}
        \sum_{k \in N} u_{jk} &= 1 \quad \forall j \in N \label{eq:eq_1} \\
        \sum_{j \in N} u_{jk} &= 1 \quad \forall k \in N \label{eq:eq_2} \\
        c_1 &= \sum_{j \in N} p_j u_{j1} \label{eq:eq_3} \\
        c_k &\geq c_{k-1} + \sum_{j \in N} p_j u_{jk} \quad \forall k \geq 2 \label{eq:eq_4} \\
        C_j &\geq c_k - M(1 - u_{jk}) \quad \forall j, k \in N \label{eq:eq_5} \\
        T_j &\geq C_j - d_j \quad \forall j \in N \label{eq:eq_6} \\
        T_j, C_j, c_k &\geq 0 \quad \forall j, k \in N \label{eq:eq_7} \\
        u_{jk} &\in \{0, 1\} \quad \forall j, k \in N \label{eq:eq_8}
        \end{align}
        
    \noindent The objective function \eqref{eq:eq_0} minimizes total tardiness. Constraints \eqref{eq:eq_1}–\eqref{eq:eq_2} ensure each job is assigned to exactly one position and each position to one job. Constraints \eqref{eq:eq_3}–\eqref{eq:eq_4} determine the positional completion times. Constraints \eqref{eq:eq_5} link positional and job-based completion times, while \eqref{eq:eq_6} defines tardiness. Finally, constraints \eqref{eq:eq_7} enforce non-negativity, and \eqref{eq:eq_8} ensure integrality.






\subsection{Valid Inequalities}

To strengthen the MIP formulation and improve the quality of the lower bounds, \cite{Keha2009} propose a class of valid inequalities. These inequalities enhance the formulation by providing tighter estimates for the completion time \(C_j\) of each job \(j\), based on the minimum amount of processing that must precede it given its position in the schedule.

Note that \(p_{jk}\) gives the minimum value the completion time of the
job at position k 1 can take given that job j is at position k. Specifically, the inequality relies on the term \(\pi_{jk}\), which represents the minimum completion time of job in position \(k-1\) that must occur given job \(j\) is assigned to position \(k\). This term accounts for the cumulative processing time of other jobs that could appear earlier in the sequence, and varies depending on the relative position of job \(j\).

The value of \(\pi_{jk}\) is defined as:
\[
\pi_{jk} = 
\begin{cases} 
\sum_{l=1}^{k-1} p_l, & \text{if } k \leq j \\[6pt]
\sum_{l=1}^{j-1} p_l + \sum_{l=j+1}^{k} p_l, & \text{if } k > j
\end{cases}
\]

This piecewise definition accounts for the relative ordering of job \(j\) and the position \(k\) to which it may be assigned. When \(k \leq j\), the minimum preceding workload is given by the cumulative processing time of the jobs scheduled in positions 1 through \(k-1\), all of which precede job \(j\). In contrast, when \(k > j\), job \(j\) would appear later in the sequence, and the cumulative processing time excludes \(p_j\), summing instead the processing times of jobs occupying positions 1 through \(j-1\) and \(j+1\) through \(k\).

The resulting valid inequality, applied to all jobs \(j \in N\), is:
\[
C_j \geq p_j + \sum_{k=2}^{n} \pi_{jk} u_{jk}. \label{eq:eq_11}
\]

This constraint ensures that the completion time of job \(j\) is at least its own processing time plus the minimum estimated time needed to process preceding jobs, depending on the position \(k\) to which job \(j\) is assigned.
If the job $j$ is at position $k > 1$, then $u_{jk}$ becomes $1$, and Equation~\ref{eq:eq_11} becomes $C_j \geq p_j + \pi_{jk}$, which is consistent with the definition of $\pi_{jk}$s.

This inequality is particularly useful when jobs are sorted in non-increasing order of processing times, as it reinforces realistic lower bounds on completion times for later jobs in the sequence. Incorporating this valid inequality helps reduce the feasible region and improves solver performance, especially on larger instances.

\subsection{Dynamic Programming (DP) Formulation}
\label{subsec:DP}
We utilize the following dynamic programming (DP) formulation to minimize total tardiness by recursively building schedules. 

\textbf{Initial Conditions:} For each job \(k \in N\), the tardiness function is:
\begin{align}
    h_k(C_k) &= \max \{0, C_k - d_k\}, \quad \forall k \in N, \label{eq:eq_12}
\end{align}
where \(d_k\) is the due date of job \(k\). The value function for scheduling only job \(k\) is:
\begin{align}
    V(\{k\}) &= h_k(C_k). \label{eq:eq_13}
\end{align}

\textbf{Recursive Relations:} For a set \(J \subseteq N\), the tardiness function for the last job \(k\) in the sequence is:
\begin{align}
    h_k(C(J)) &= \max \left\{0, \sum_{j \in J} p_j - d_k \right\}, \quad C(J) = \sum_{j \in J} p_j. \label{eq:eq_14}
\end{align}
The value function \(V(J)\) recursively selects the optimal sequence by minimizing total tardiness:
\begin{align}
    V(J) &= \min_{j \in J} \left( V(J \setminus \{j\}) + h_j(C(J)) \right). \label{eq:eq_15}
\end{align}

    \section{Methods}

    In this section, we present the methodology of our study. The section starts with the description of our framework to discover heuristics for the single machine total tardiness scheduling (SMTT) problem and continues with exact methodological setup and concludes with the proposed algorithms.

\subsection{Scientific Discovery Using LLMs}
\label{subsec:pipeline}

In this subsection, we describe the heuristic discovery pipeline that is used throughout the study. The methodology employed in this study involves developing novel heuristics by leveraging the capabilities of LLMs. The problem under concern is an NP-hard problem; by nature, NP-hard problems are ‘hard-to-solve’ to optimality but the solutions to these problems are ‘easy-to-evaluate’. In other words, while the feasibility or objective value of a proposed solution to an NP-hard problem can often be evaluated in polynomial time, finding an optimal solution is computationally intractable in the general case \citep{nemhauser1988integer, garey1979computers}. The framework proposed to discover new heuristic rules for an NP-Hard Mixed Integer Program is inspired by the FunSearch method developed by Google DeepMind (\cite{funsearch}).

Although LLMs have advanced significantly, they alone remain insufficient to push the state-of-the-art forward when addressing NP-hard problems, where finding optimal solutions is computationally expensive. In such cases, creative heuristics are needed to navigate large solution spaces efficiently, and while LLMs can assist in generating new ideas, they must be combined with traditional algorithms or frameworks (e.g., heuristics) to achieve practical improvements.

LLMs, while powerful, often produce inaccurate or suboptimal outputs, commonly known as ``hallucinations,'' when applied to scientific discovery tasks. To enhance their reliability and problem-solving capability, it is essential to embed LLMs within a structured framework that enables repetitive and incremental innovation and validation. To extend the capabilities of LLMs, our methodology combines the flexibility, speed, and power of LLMs with evolutionary algorithms. To this extent, an island-based evolutionary approach is used to protect the LLM from ‘confabulations’ and ineffective ideas. While our method draws inspiration from FunSearch, it is designed with several key adaptations such as specification of the problem and the rigorous evaluation process to meet the structural and computational needs of operations research (OR) problems, particularly MIP-based scheduling. We reinforce our specification with `prompt engineering' to inform LLM about the problem structure. Compared to FunSearch, our approach introduces rigorous testing against stronger baseline algorithms and also the optimal solutions of the problems which is an innovation rooted in the combinatorial optimization paradigm. We test our algorithms against the optimal tardiness values while \cite{funsearch} compares their bin-packing algorithms' performance against simple heuristic rules' performance such as best-fit and first-fit.

\begin{figure}
\centering
\includegraphics[width=1.0\linewidth]{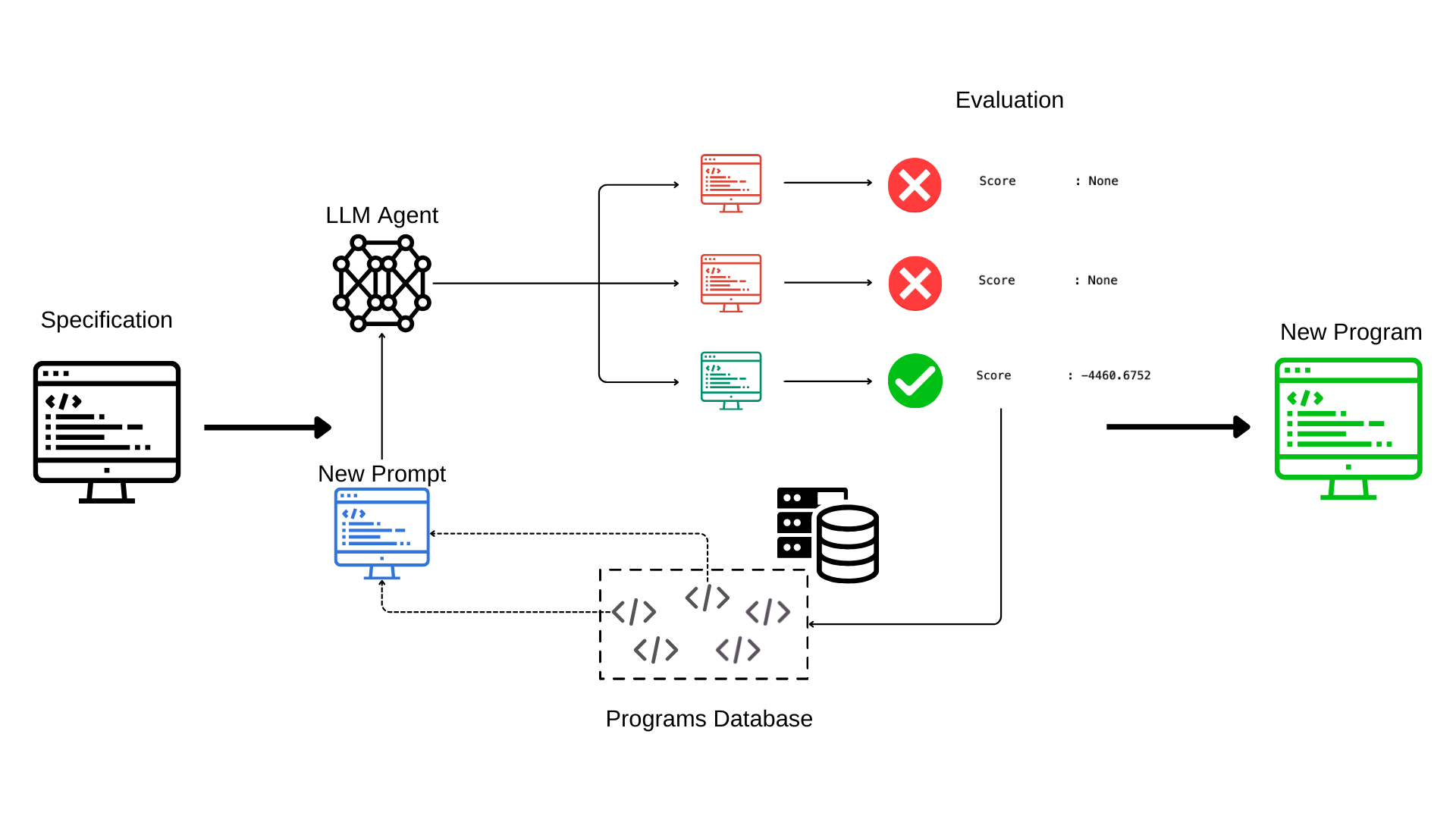}
\caption{Illustration of the heuristic discovery pipeline (inspired by \cite{funsearch}). Figure created by the authors.}
\label{fig:pipeline_figure}
\end{figure}

The FunSearch pipeline is composed of several counterparts, as demonstrated in Figure \ref{fig:pipeline_figure}:
\begin{itemize}
    \item \textbf{The Specification} is a concise code snippet that translates the mathematical problem into computer code, comprising an ‘assignment’ and an ‘evaluate’ function. It also includes a short description of the problem and what is expected from the LLM Agent to `prompt engineer' the and get more relevant outputs.

    \item \textbf{The LLM Agent} refines the programs provided through prompts. It serves as a ‘mutation operator’ in an evolutionary process, developing high-performing heuristics. For this, a pre-trained — or ‘frozen’ — LLM is employed, with its parameters kept constant across iterations, solely used for inference to generate new code variants.

    \item \textbf{The Evaluator} assesses the programs generated by the LLM Agent by assigning scores. Solutions are tested on datasets to verify feasibility and performance using the specified ‘evaluate’ function. Feasibility is enforced through soft constraints: schedules containing duplicate jobs or missing jobs are penalized (see Section \ref{subsec:specification}). This component is critical for guiding the LLM, preventing ‘confabulations’ (false solutions), and steering it toward productive improvements by measuring previous outputs’ effectiveness.

    \item \textbf{The Programs Database} stores ‘correct’ programs produced by the LLM Agent. Feasible solutions are logged, sampled for new prompts, and reused in subsequent iterations to enhance performance.
\end{itemize}

Our pipeline builds on the essence of \cite{funsearch} but is tailored specifically to solve a mixed-integer program with an application to the single machine total tardiness scheduling problem, focusing on efficiency and scalability. We introduce a single-threaded resource-efficient approach, in contrast to the multi-threaded, resource-intensive setup in \cite{funsearch}, which relies on multiple samplers and evaluators to improve capacity. Despite the simpler configuration, our method maintains comparable effectiveness with a single sampler and evaluator, making it well-suited for practical applications on MIPs. Additionally, our specification is customized for the scheduling domain to address its unique constraints and objectives, which differ significantly from the online bin-packing problems tackled in their framework. Moreover, our methodology includes testing and comparing the resulting algorithms to stronger baseline algorithms, emphasizing generalization across problem sizes. We test our algorithms against the optimal tardiness values and complicated algorithms in the domain while \cite{funsearch} compares their bin-packing algorithms’ performance against the objective values of simple heuristic rules such as best-fit and first-fit.

We opted for a single-threaded implementation primarily for practical and reproducibility reasons. This setup enables easy deployment on standard computing environments without the need for large-scale parallel infrastructure. While a multi-threaded design could accelerate sampling and evaluation, it introduces additional complexity in managing synchronization, load balancing, and memory usage across threads. In our experiments, the single-threaded configuration was sufficient to achieve meaningful results within reasonable time and resource constraints demonstrating that an effectively configured single GPU can be enough to discover heuristics.

The pipeline begins with the provided problem specification. The LLM agent uses the ‘assignment’ function as input to generate new programs (e.g., heuristics) aimed at solving the problem. These programs are evaluated through the ‘evaluate’ function. Programs that produce valid results are stored in the ‘programs database’ and ranked based on their performance metrics. 
This approach departs from classical genetic programming for combinatorial optimization problems. Rather than evolving a population of candidate solutions directly, our method maintains a population of solution functions (or heuristics) and iteratively builds new functions on top of the existing ones. Programs stored in the database are further refined using the ‘best-shot prompting’ technique, which combines multiple well-performing functions into a single prompt through sampling. In our experiments, we found that sampling two programs per prompt provides a good trade-off between the LLM’s creativity and computational efficiency. The enhanced program is then reintroduced to the LLM for further iterations. It should be noted that at each iteration, LLM receives the following prompt as a string alongside the program: `Find the mathematical heuristic function for the single machine scheduling problem that returns indices of assigned jobs, given data on processing times, due dates of jobs. Note that each job is assigned to the machine exactly once (i.e., no job is left unassigned or have multiple assignment). Note that the due dates and the processing times should not be manipulated.' This is an example of `prompt engineering,' a common industry practice, in which we inform the LLM Agent about the problem to guide better solutions.

As shown in Figure \ref{fig:pipeline_figure}, an island-based evolution method is employed to refine the programs over multiple iterations. For the experiments, 10 islands are used, with a random island selected at each step to register the next generated function. The island selection follows a discrete uniform distribution and the prompt is constructed using functions of the selected island. Every 14,400 seconds (4 hours) of runtime, the weaker half of the islands are reset. The performance of the island is determined by the best performing function within each island. When an island is reset, it is re-initialized using a random, non-reset island — selected via discrete uniform distribution — as its ‘founder’ island. The best function of the founder island is transferred to the newly reset island, allowing the population to evolve iteratively until the next reset.

The island-based strategy supports exploration and mitigates premature convergence by maintaining diverse sub-populations of heuristics. Each island evolves independently, which increases the chance of discovering novel yet feasible program variants. The periodic reset mechanism allows stagnant islands to benefit from stronger ones without collapsing global diversity. This design is particularly advantageous for OR problems where multiple promising regions may exist in the solution space, and preserving algorithmic diversity is key to finding robust heuristics.

\begin{figure}
\centering
\includegraphics[width=1.0\linewidth]{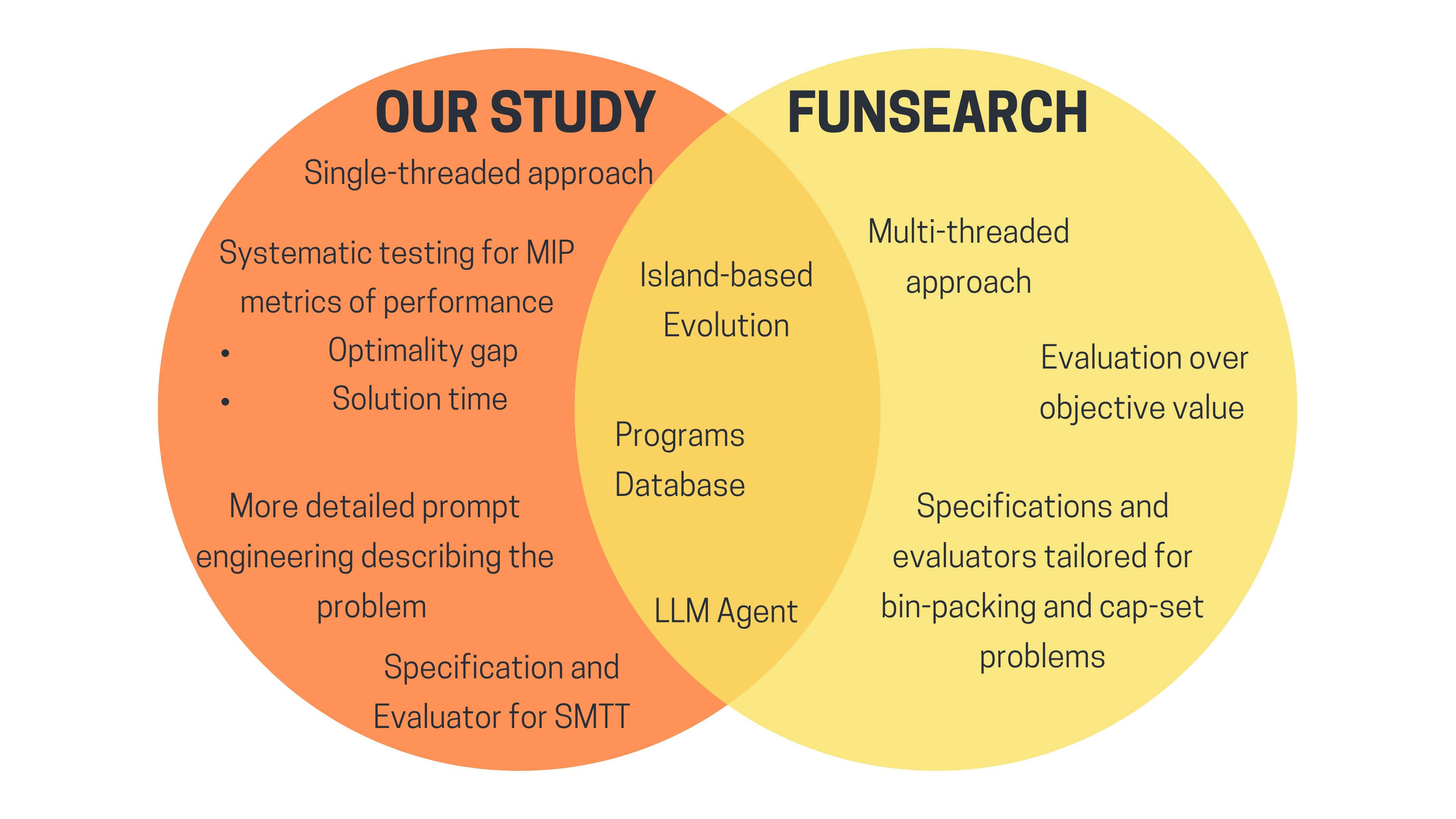}
\caption{Comparison of our study and FunSearch (\cite{funsearch})}
\label{fig:comparison_figure}
\end{figure}

To sum up, Figure \ref{fig:comparison_figure} provides an overview for the comparison of our work and FunSearch (\cite{funsearch}). 
Although we share the main framework with FunSearch (Island-based Evolution, Programs Database structure and utilizing an LLM Agent), FunSearch applies a resource-intensive, multi-threaded approach whereas we opt for a single threaded approach to demonstrate the efficacy of the method with less resources. While FunSearch evaluate their developed programs using the objective values, we test rigorously against the optimal solutions and compare solution times among algorithms. Lastly, FunSearch utilizes specifications and evaluators tailored for the bin-packing and cap-set problems while we use a specification and an evaluator geared towards SMTT and back up with prompt engineering methodology.

\subsection{FunSearch Heuristic Discovery Process for the Single-Machine Total Tardiness (SMTT) Problem}

    \label{subsec:specification}

    In this subsection, we present the problem specification that initiates the algorithm discovery pipeline. A tailored specification is designed specifically for discovering new heuristics for the single-machine total tardiness scheduling problem. This specification translates the problem and its constraints into computer code, enabling a structured interaction with the LLM. By implementing this specification, we create a sandbox environment that allows the LLM to generate and improve `assignment' functions effectively.
    
   The specification consists of two main components: the `assignment' and `evaluate' functions. The `assignment' function provides an initial solution method, which serves as the starting point for the pipeline to modify and generate new `assignment' functions. Through these modifications, the LLM agent explores potential improvements, storing `correct' functions in the `programs database'. A function is only stored in the `programs database' if it produces valid results within the specified time limit. The systematic evaluator ensures the correctness of the functions generated. It verifies the feasibility of the SMTT schedules produced by the `assignment' function and assesses their performance by calculating total tardiness over the provided dataset. The `evaluate' function guides the LLM agent by identifying areas for improvement and steering the search toward more effective solutions. 
    
    During preliminary trials, some programs generated by the LLM exhibited undesirable behavior. Specifically, some programs assigned only a subset of jobs or duplicated jobs to minimize tardiness. Others manipulated the processing times and due dates, effectively altering the problem constraints to achieve better results than the optimal solution. To mitigate these irregularities, we designed the problem specification to penalize incomplete or manipulated solutions. Schedules that omit jobs or assign the same job multiple times, utilizing scheduling flexibility, receive very high penalty scores. Additionally, the input data is treated as immutable, preventing the `assignment' function from altering it. These soft constraints strike a balance between maintaining the flexibility of the LLM and reducing unintended behaviors. See Figure \ref{fig:evaluate_figure} for details on our `evaluate' function.

For the `assignment' function, a prompt describing the problem and several well-known heuristic rules from the literature alongside  are used as starting points, yielding various results. The EDD, SPT, and MDD rules are implemented as `assignment' functions, each running in separate pipelines to allow the LLM to modify them.  The PSK and Pannerselvam heuristics were not used as `assignment' functions due to two key considerations: (1) more complex `assignment' functions reduce the likelihood of generating valid programs, and (2) they increase iteration runtimes. Preliminary trials with the SPT pipeline showed poor performance, so it was excluded from the results section. However, the EDD and MDD `assignment' functions led to the development of the EDD Challenger (EDDC) and MDD Challenger (MDDC) functions, respectively. Figure \ref{fig:assignment_figure} provides an example of the EDD-based `assignment' function. To guide the LLM towards generating more relevant results, a docstring is included at the beginning of the code. This docstring introduces the mathematical problem, defines the inputs, outputs, and their data types (see Figure \ref{fig:docstring_figure}). Additionally, two instructional sentences are added to the docstring to ensure that the LLM aligns with the problem's constraints during the solution process.

\subsection{The Experimental Setup and Training}

In this subsection, we present the setup and details of our experiments. In our study, the Mixtral 8x7B instruction model, which is an open source LLM, is used as the backbone LLM, while \cite{funsearch} employ Codey. Selecting a backbone LLM involves an inherent trade-off between sample quality and inference speed. Both Mixtral and Codey are positioned similarly in this spectrum, offering fast inference speeds. Furthermore, the performance of our experiments does not depend heavily on the exact choice of LLM (\cite{funsearch}). The pipeline is deployed on Virginia Tech's Advanced Research Computing Center clusters Nvidia A100 GPUs with 80GB memory and AMD EPYC 7742 chip with Zen 2 architecture for inference. Although the original study by \cite{funsearch} employed multiple parallel evaluators and samplers, we adopted a single-threaded version due to resource constraints. An environment is constructed using Python version 3.11 using Transformers (4.43.3), Tensorboard (2.17.0) and PyTorch (2.3.0) libraries for the LLM inference and Gurobi 11.0.1 for optimization models.

The training (the heuristic discovery pipeline) begins with the initial specification and tests the generated functions on a dataset of 10,000 problem instances, each with a 25-job problem size. Details of the data generation schema are provided in Section \ref{s:datagen}. The choice of problem size in the training dataset is a critical decision, as the generalizability of the resulting algorithms improves with larger problem instances. However, larger problem instances require more time for evaluation, which reduces the number of iterations the pipeline can complete within a given time frame. Through preliminary trials, we determined that the size of a 25-job problem strikes a balance between the evaluation time and the generalizability of the result. The experiments are conducted over 72 hours on the specified computing clusters, with a maximum limit of 10,000 iterations. The pipeline halts when either condition is met. Although our single-threaded approach involves fewer computational resources compared to previous studies, we successfully develop algorithms that advance the state-of-the-art, as detailed in Section \ref{subsec:algorithms}. 

While the heuristic generation phase requires a one-time computational investment of up to 72 hours, this process yields high-performing and reusable algorithms that can be deployed repeatedly across a wide range of instances with negligible additional cost. Once discovered, these heuristics are significantly faster than traditional exact methods and even outperform well-known heuristic baselines in solving NP-hard SMTT problems, as shown in Section \ref{Performance Analysis of Discovered Heuristics}. Thus, the up-front training effort results in substantial long-term computational savings, making the approach both scalable and practical for real-world applications.

    \begin{figure}[h!]
    \centering
    \includegraphics[width=1.0\linewidth]{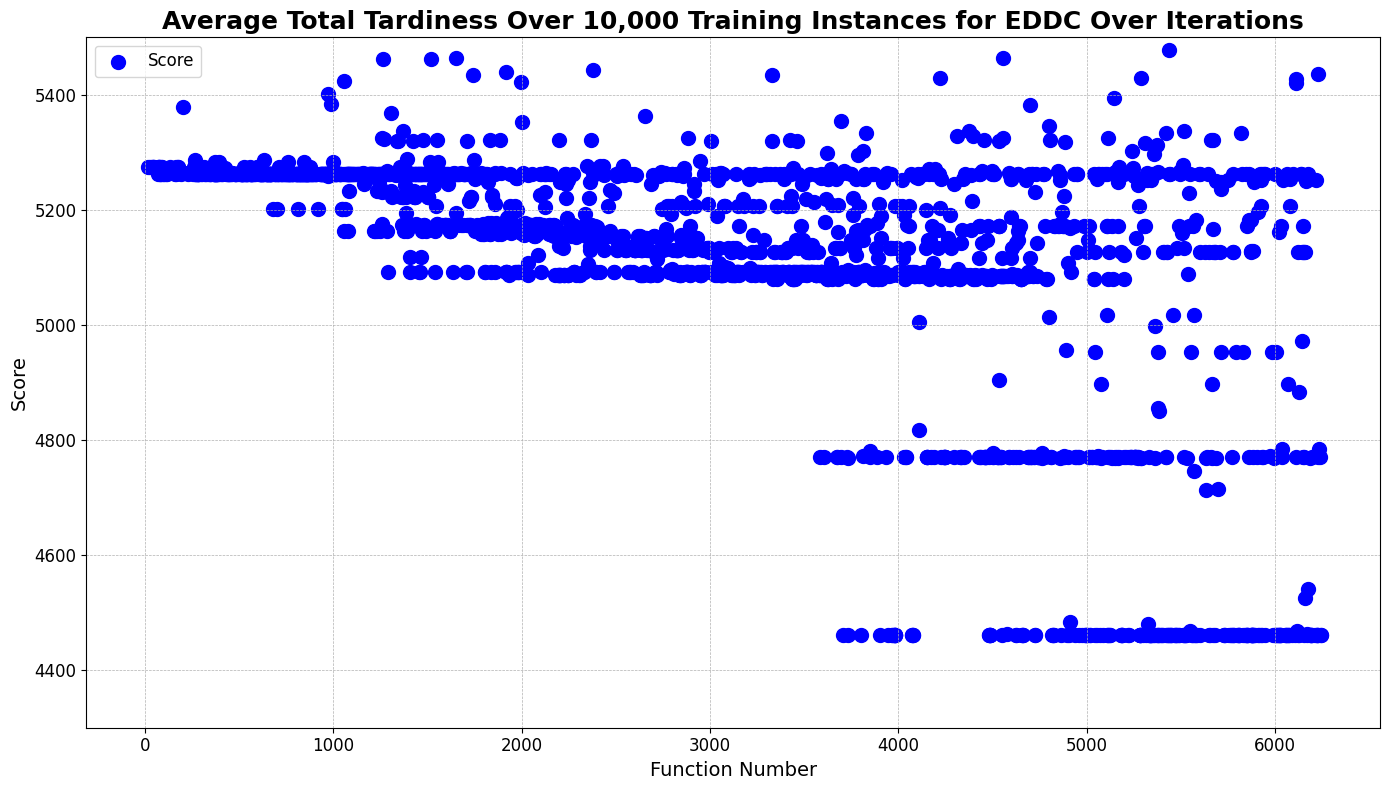}
    \caption{Average Total Tardiness Throughout the Training for EDDC Over Iterations}
    \label{fig:eddc_overiter}
    \end{figure}  

\begin{figure}[h!]
    \centering
    \includegraphics[width=1.0\linewidth]{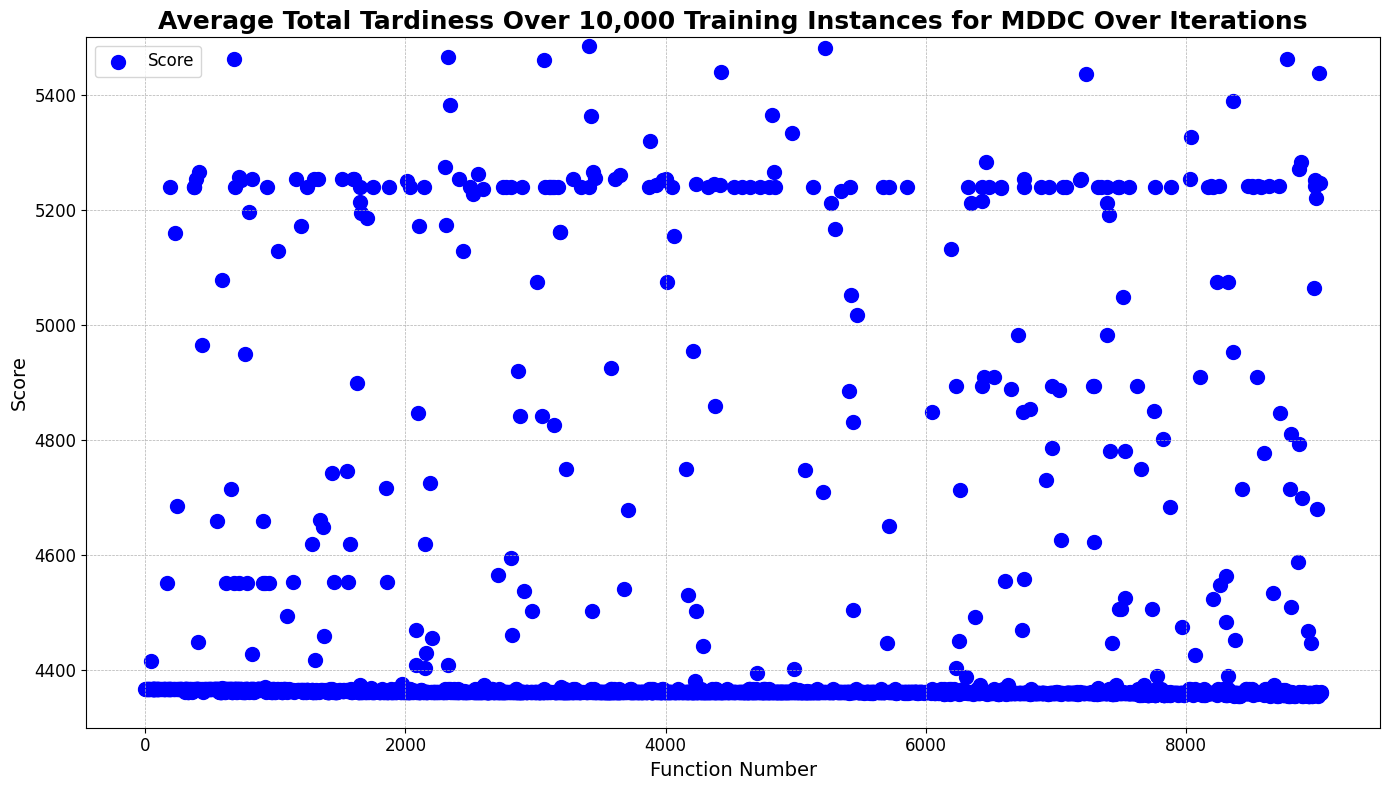}
    \caption{Average Total Tardiness Throughout the Training for MDDC Over Iterations}
    \label{fig:mddc_overiter}
    \end{figure}

      Figures \ref{fig:eddc_overiter} and \ref{fig:mddc_overiter} show the average tardiness values of the EDDC and MDDC algorithms throughout the training process, respectively. The objective of the pipeline is to minimize the average total tardiness of the training dataset. The EDDC algorithm starts with an initial score around 6400 and gradually improves to 4680 by the end of training. The plateau-like structure of the plot indicates that the algorithm evolves incrementally, building on previous solutions over time. It should also be noted that the process includes generation of some inferior programs which can be observed as the `outlier' dots in Figures \ref{fig:eddc_overiter} and \ref{fig:mddc_overiter}. These programs are not discarded immediately and kept in the database, although they are not preferred as much as the ones with a lower tardiness in the best-shot prompting. In contrast, the MDDC algorithm begins with a strong initial score of 4362 and enhances it slightly to 4355 through the application of additional arithmetic operations on top of the original MDD algorithm. While this change may seem minor, it reflects a meaningful improvement, advancing the state of the art, as discussed in detail in Section \ref{s:numerical}.

    \subsection{Two New Algorithms Discovered by LLMs}
    \label{subsec:algorithms} 
  
In this subsection, we introduce the algorithms developed through the scientific discovery pipeline, which incorporates the problem specifications, the heuristic discovery process, and the training framework as detailed in Sections \ref{subsec:pipeline} and \ref{subsec:specification}.

\begin{algorithm}
    \caption{EDD Challenger (EDDC) Algorithm: A local search routine is integrated into the original EDD algorithm. This routine evaluates the feasibility of avoiding delays specifically for the last two jobs in the schedule. Differences with respect to EDD are highlighted in bold.}
    \label{alg:EDDC}
    \begin{algorithmic}[1]
     
    \State {Input:} Arrays of processing times $\boldsymbol{p}$ and due dates $\boldsymbol{d}$ for each job
    \State {Output:} schedule ${S}$
    \Procedure{EDDC}{${p}$, ${d}$}
    \State Initialize $S$ by sorting the jobs lexicographically based on $d$ and $p$
    \For{$\boldsymbol{i \gets 1}$ to $\boldsymbol{|S|}$}
        \State $\boldsymbol{j \gets i}$
        \While{$\boldsymbol{j > 0}$ and $\boldsymbol{d[S[j]] < d[S[j - 1]]}$}
            \State $\boldsymbol{T_{\text{completion}} \gets \sum p[S[:j]]}$ 
            \If{$\boldsymbol{j = |S|}$ and $\boldsymbol{d[S[j - 1]] > T_{\text{completion}}}$} 
                \State Swap $\boldsymbol{S[j]}$ and $\boldsymbol{S[j - 1]}$
            \EndIf
            \State $\boldsymbol{j \gets j - 1}$
        \EndWhile
    \EndFor
    \State \Return ${S}$ 
    \EndProcedure
\end{algorithmic}
\end{algorithm}


The EDD Challenger (EDDC) (Algorithm \ref{alg:EDDC}) is based on the EDD strategy and includes a local search routine that extends beyond the simple EDD assignment function. After about 68 hours and 10,000 iterations, the algorithm emerged as the top performer during its 6,841\textsuperscript{st} iteration. In our approach, the most recent algorithm found at the end of the iterations is not always the best; instead, our methodology maintains a database of all generated algorithms, allowing continuous refinement and selection of the best-performing option.

 The EDDC Algorithm (Algorithm \ref{alg:EDDC}) improves the one-shot scoring logic of the EDD rule with a local search routine. The algorithm starts by sorting the jobs lexicographically based on due dates and processing times, just like EDD. Moreover, it incorporates a search routine utilized by a `for loop' that iterates over each job in the initial schedule, to check some conditions to make changes. EDDC first checks if the due date of the second last job in the schedule's due date is later than the last job or not; if so, $T_{\text{completion}}$ is calculated which is the completion time of the currently scheduled jobs. The final condition to check is whether the second last job's due date is larger than $T_{\text{completion}}$ or not, if yes, this means we can delay the second last job, and last job and second last job are swapped. This body of rules can be interpreted as a simple optimality check; however, it increases the effectiveness of EDD, as is mentioned in Section \ref{s:numerical}. The original EDD algorithm is easy to implement, making it a less complex option but with a higher optimality gap, whereas EDDC is a more complex algorithm that achieves a lower optimality gap. The LLM innovations in the original EDD algorithm are highlighted in bold within the pseudocode (Algorithm \ref{alg:EDDC}).
    
The MDD Challenger (MDDC) algorithm (Algorithm \ref{alg:MDDChallenger}) is developed using a specification that incorporates the MDD rule as its `assignment' function. The pipeline ran for 72 hours, completing 9,094 iterations, indicating that the sampling times were longer compared to the EDD pipeline. We attribute this increase to the greater complexity and length of the MDD specification. Although several algorithms that performed well earlier during the discovery process were generated, MDDC emerged as the best algorithm in the \nth{8398} iteration.

    \begin{algorithm}
    \caption{MDD Challenger (MDDC) Algorithm: Incorporates dynamic scoring adjustments through $\rho$, $\theta$, and $\sigma$ to enhance scheduling decisions. Differences with respect to MDD are highlighted in bold.}\label{alg:MDDChallenger}
    \begin{algorithmic}
        \State \textbf{Input:} Arrays of processing times $p$ and due dates $d$ for each job
        \State \textbf{Output:} schedule $S$
        \Procedure{MDDC}{$p$, $d$}
            \State Initialize $current\_time$ as 0,  $S$ as an empty set, unscheduled jobs $U$ as the set of all jobs 
            \State Sort $U$ based on Shortest Processing Time (SPT) Rule
            \State Initialize $p_U$ and $d_U$ as the processing times and due dates of all unscheduled jobs

            \While{$|U| \neq 0$}
                \State $\mu \gets \max\left(p_U \, \mathbf{\times \, \textbf{1.1}} + current\_time, d_U\right)$
                \textbf{\State $\boldsymbol{\rho \gets \min\left(\frac{p_U}{current\_time + \max(p_U)}, 1\right)}$
                \State $\boldsymbol{\theta \gets \frac{\rho^2}{1 + \rho^2}}$
                \State $\boldsymbol{\mu \gets \mu \times (1 + \theta)}$
                \State $\boldsymbol{\sigma \gets \frac{p_U}{current\_time + \sum(p_U) / |U|}}$
                \State $\boldsymbol{\mu \gets \mu + \sigma}$}
                \textbf{\State Find the job with the minimum $\boldsymbol{\mu}$ score and schedule it at the last position }
                \textbf{\State Update $\boldsymbol{U}$, $\boldsymbol{S}$, $\boldsymbol{current\_time}$, $\boldsymbol{p_U}$, and $\boldsymbol{d_U}$}
            \EndWhile
            \State return $S$
        \EndProcedure
    \end{algorithmic}
\end{algorithm}

MDDC adopts an iterative scheduling approach based on MDD's initial structure with a more complex score calculation mechanism. It sequences jobs one by one across multiple iterations using a while loop, recalculating scores for unscheduled jobs at each iteration until there are no unscheduled jobs left. This design closely mirrors the logic of the MDD heuristic but with additional terms that extend the original MDD formulation. The algorithm begins by initializing $current\_time$ to 0, incrementing it by the processing time of each newly scheduled job. The set $U$ contains all unscheduled jobs (initially, every job) and is sorted by the Shortest Processing Time (SPT) rule. The set $S$ stores scheduled jobs and starts as an empty set. Arrays containing the processing times and due dates of jobs are initialized with values for all jobs and are updated after each assignment to reflect only unscheduled jobs. The final score $\mu$ is computed through algebraic operations that incorporate several sub-scores. Initially, $\mu$ is set using a weighted term similar to the original MDD score, \(\max(p_U + \text{current\_time}, d_U)\). Next, the sub-score $\rho$ is computed as the minimum of one and a scaled processing time. However, the use of the minimum operator is redundant, as \(\frac{p_U}{\max(p_U)}\) will not exceed one, especially since the denominator also includes the current time. This reflects a limitation of the LLM, which, while capable of generating new solutions, lacks a deep understanding of the domain or the logic of the problem.

The sub-score $\theta$ is then calculated and incorporated into the main score, $\mu$. The final sub-score, $\sigma$, is obtained by dividing the processing time of each job by the sum of the current time and the average processing time across unscheduled jobs. This value is added to $\mu$ to compute the ultimate score. Since $p_U$ and $d_U$ represent the processing times and due dates of unscheduled jobs, the scores are recalculated for each unscheduled job in every iteration. The job with the minimum $\mu$ score is scheduled next, and the process repeats until all jobs are scheduled. The LLM’s improvements on the original MDD algorithm are highlighted in bold within the pseudocode (Algorithm \ref{alg:MDDChallenger}).

Our approach leverages the flexibility of LLMs to explore a wide range of options. The LLM agent systematically tests thousands of arithmetic operations and ideas from other disciplines to derive effective rules. Thousands of iterations are carried out, with incremental changes made based on the previously generated programs, resulting in a final algorithm built on cumulative improvements. This approach offers both advantages and challenges. On the positive side, the heuristic rule introduces several novel operations that are not found in existing heuristics from the literature. However, the methodology can be difficult to interpret and may not always align with the logical principles of the scheduling domain since the sub-routines or sub-scores within the algorithm are not always connected naturally.

The scores observed in MDDC resemble regressive terms in statistics, reflecting the underlying assumptions and working principles of the pipeline. Rather than designing a solution methodology from scratch using human expertise and domain knowledge, the LLM makes small incremental changes to an initial heuristic code at each iteration and evaluates their effectiveness. If a change improves the performance of the heuristic, it is prioritized among the generated programs in the database and used as a future prompt to guide subsequent iterations. For example, during the second experiment, the LLM agent modified the initial score calculation of the MDD rule from $\max(p_U + current\_time, d_U)$ to $\max(p_U \times 1.1 + current\_time, d_U)$. This modification, introduced in the \nth{304} iteration, reduced the average total tardiness and became the foundation of many subsequent programs. In this way, beneficial small changes are retained in the program database and accumulated over time to produce a better-performing algorithm.

For a 6-job example with processing times [10, 11, 10, 10, 11, 10] and due dates [15, 11, 13, 11, 12, 11], the MDDC sequences the jobs in the order of 3, 5, 2, 0, 1, and 4, while the original MDD rule sequences them as 1, 0, 2, 3, 5, 4. The schedule generated by MDDC results in 141 time units of tardiness, which matches the optimal value for this problem instance. In contrast, the MDD rule produces a schedule with 144 time units of tardiness. Although the original MDD heuristic achieves a near-optimal solution, the enhancements introduced by the LLM enable the MDDC to reach the optimal solution, beating the original MDD heuristic in this instance.

Overall, the reliance of the heuristics discovery pipeline on LLMs reduces the explainability of the resulting algorithms. At each iteration, the LLM introduces changes inspired by various domains, implementing these modifications incrementally. As a result, the heuristics evolve through a series of additive improvements, making their underlying logic increasingly difficult to interpret. It is important to note that randomness plays a significant role in these experiments, as the outputs of LLMs are inherently non-deterministic. Furthermore, the cumulative use of LLM-generated prompts introduces compounding stochasticity across iterations, contributing to variability in the heuristic discovery process.



The logical schema generated by EDDC shows that our pipeline can develop rules to check conditions for lower tardiness values, resembling those proposed by \cite{Emmons1969}. This represents a significant advancement, as the pipeline was able to discover more complex decision rules than \cite{funsearch}, even though it did not yield the best-performing algorithm (see Section \ref{Performance Analysis of Discovered Heuristics}).
However, we also observe that the pipeline retains redundant manipulations in its program database and lacks a holistic view of the problem. These redundancies, while occasionally useful, can obscure the core logic of the final heuristic.  Incorporating human expertise could help refine this mechanism and lead to new conditions for tardiness reduction. Moreover, the rediscovered conditions highlight that, despite limited transparency, the pipeline can recover classic scheduling principles. This suggests an emergent form of explainability, with rule patterns that echo established theory. Future efforts to improve explainability could include incorporating interpretable surrogate models, such as decision trees that approximate the heuristic’s behavior and reveal key decision rules. Constraint-based pruning could be used to filter out redundant or logically inconsistent rules by enforcing known scheduling principles (e.g., removing jobs with dominated priority conditions). Additionally, LLM-driven changes could be summarized in human-readable formats, such as visual diffs or annotated logs that explain why each modification was introduced. These approaches can bridge the gap between the generative creativity of LLMs and the transparency required for trustworthy, human-in-the-loop decision-making. In the next section, we present the experimental evaluation of the discovered heuristics across different problem sizes.

    \section{Experimental Design and Results} \label{s:numerical}

    In this section, we first present the data generation schema and the experimental design for this study and then present the computational results. 

    \subsection{Data Generation and Experimental Design} \label{s:datagen}
In this study, we employ a data generation schema inspired by \cite{Potts1982}. Processing times are uniformly distributed between 1 and 100, and due dates are governed by two parameters: the relative due date range ($RDD \sim \{0.2, 0.4, 0.6, 0.8, 1.0\}$) and the average tardiness factor ($TF \sim \{0.2, 0.4, 0.6, 0.8\}$). TF = 1.0 instances are excluded from the test, as they are considered easier instances (\cite{Potts1982}), allowing us to focus on more challenging cases where the performance of the optimality gap is observed to be more significant. However, TF = 1.0 instances are included during training to provide a more diverse dataset, promoting better generalization of the algorithms. The due dates are generated using a uniform distribution $U\left[ P(1-TF-\frac{RDD}{2}), P(1-TF+\frac{RDD}{2})\right]$, where $P = \sum_{i=1}^{N} p_i$. 

For testing, we use instances with $\left| N \right|$ = $20$, $100$, $200$, and $500$. We generate 800 instances for \(\left| N \right| = 20\) using two different distributions for processing times. The first distribution is the uniform distribution between 1 and 100, as mentioned earlier. The second distribution is normal, with a mean of 60 time units and a standard deviation of 20 units. The second dataset aims to evaluate the algorithms' performance on out-of-sample instances. Both problem datasets with $\left| N \right|$ = $20$ are solved using the Dynamic Programming approach presented in Section \ref{subsec:DP}. The instances with $\left| N \right|$ = $100$, $200$, and $500$, each comprising 200 problems, along with their solutions obtained using the Branch and Memorize algorithm, are sourced from \cite{vtkindt}. It should be noted that the training process is completed using a 25-job $(|N|= 25)$ dataset with 10,000 instances to advance the heuristics and later on, resulting algorithms are applied to larger problems up to 500-job problems without further training.

\subsection{Computational Results} \label{s:numerical}

 In this subsection, we present the computational results for the methodologies presented so far. While Table \ref{table:VIvsDP} compares the run times of exact solution methods, Tables \ref{table:optimality20}, \ref{table:optimality_outofsample}, \ref{table:optimality_100}, \ref{table:optimality_200_jobs} and \ref{table:optimality_500_jobs} demonstrate the comparison of the discovered heuristics with the current heuristics in the literature over a select subset of instances while giving the overall average results. Then, Table \ref{table:optimality_scale} compares the optimality gaps of the algorithms: MDD, MDDC, and their augmented versions; and lastly, Table \ref{table:solution_times_scale} compares the solution times in CPU seconds. For clarity, Tables \ref{table:optimality20} to \ref{table:optimality_500_jobs} present six different combinations of problem parameters in each row, illustrating the performance of various solution methods on problem instances of varying difficulty, while the final row reports the average performance across the entire dataset of 800 problem instances, excluding zero-optimal cases.


 The optimality gap of a heuristic approach is computed as the percentage difference between the total tardiness value of the heuristic solution and the optimal tardiness value found by an exact approach (DP or Branch and Memorize of \cite{shang2021}). This gap reflects how closely the heuristic approximates the optimal solution of the MIP formulation \eqref{eq:eq_1}–\eqref{eq:eq_8}. The percent optimality gap is computed as:
\begin{equation}
\text{Optimality Gap (\%)} = 
\frac{\text{Heuristic Tardiness Value} - \text{Optimal Tardiness Value}}
{\text{Optimal Tardiness Value}} \times 100
\label{eq:opt_gap}
\end{equation}

\subsubsection{Performance Comparison of Exact Methods}

Table \ref{table:VIvsDP} presents a comparative analysis of the solution times between the SMTT MIP formulation \eqref{eq:eq_1}--\eqref{eq:eq_8}, the SMTT MIP formulation enhanced with valid inequalities (VI) \eqref{eq:eq_11}, and the dynamic programming (DP) formulation \eqref{eq:eq_12}--\eqref{eq:eq_15}. Each method is evaluated on instances characterized by three parameters: $n = |N|$, the number of jobs in the instance; TF, the Tardiness Factor; and RDD, the Relative Due Date. These parameters define the structure of the problem and allow us to evaluate performance under different scheduling conditions. The columns \textbf{timeMIP}, \textbf{timeVI}, and \textbf{timeDP} denote the average solution times in CPU seconds for the MIP and MIP with VI approaches obtained using Gurobi V11.0.1 \citep{gurobi}, and for the DP approach obtained separately using the DP formulation \eqref{eq:eq_12}--\eqref{eq:eq_15}. The terms \textbf{unsolvedMIP} and \textbf{unsolvedVI} indicate the number of instances that could not be solved within a time limit of 1 hour for the respective MIP formulations.

\begin{table}[h!]
    \caption{Computational times in CPU seconds and optimal values where $n = |N|$ represents the number of jobs}
    \label{table:VIvsDP}
    \centering
    \resizebox{\textwidth}{!}{%
    \begin{tabular}{ccccccc}
    \hline
    (n, RDD, TF) & timeMIP & timeVI & timeDP & unsolvedMIP & unsolvedVI & Optimal Value \\
    \hline
    (20, 0.2, 0.2) & 0.2 & 0.1 & 8.1 & 0 & 0 & 153 \\
    (20, 0.4, 0.4) & 1235.8 & 51.4 & 7.8 & 0 & 1 & 3910 \\
    (20, 0.6, 0.6) & 1204.4 & 1201.4 & 8.0 & 2 & 2 & 1634 \\
    (20, 0.8, 0.8) & 312.3 & 86.5 & 8.0 & 1 & 0 & 1264 \\
    \hline
    \end{tabular}
    }
\end{table}


The comparison in Table \ref{table:VIvsDP} serves as a baseline evaluation of exact approaches (MIP, VI, and DP) to highlight their computational bottlenecks and thereby motivate the development of heuristics. The table illustrates the performance trends of these exact methods under varying levels of problem difficulty. Comparisons on larger instances are omitted, as exact methods become computationally intractable once $n > 20$. Instead, larger problems (up to $n=500$) are employed as benchmark instances in Section \ref{Performance Analysis of Discovered Heuristics} to assess and compare the performance of the discovered heuristics.

    Although both MIP formulations achieve optimal solutions faster than the DP approach for easier instances with RDD and TF values of 0.2, they struggle with harder instances, where DP performs more effectively. The solution time for DP remains nearly constant within the range [7.8, 8.3] seconds, as it primarily depends on the number of nodes in the dynamic program, rather than the difficulty of the problem. However, as the number of jobs increases, the DP's solution time grows significantly. For an instance with 25 jobs, DP requires 362.5 seconds, while for 30 jobs, the memory limit is exceeded after 3 hours without a solution.

 \subsubsection{Performance Analysis of Discovered Heuristics}  
  \label{Performance Analysis of Discovered Heuristics}  

The following abbreviations are used consistently in Tables \ref{table:optimality20} to \ref{table:optimality_500_jobs}, which present results for 20, 100, 200, and 500-job problem instances, respectively, following a unified schema.

\begin{itemize}
    \item \textbf{EDD}: Percent optimality gap for the Earliest Due Date (EDD) rule
    \item \textbf{EDDC}: Percent optimality gap for the EDD Challenger (EDDC) Algorithm
    \item \textbf{MDD}: Percent optimality gap for the Modified Due Date (MDD) Heuristic
    \item \textbf{MDDC}: Percent optimality gap for the MDD Challenger (MDDC) Algorithm
    \item \textbf{PSK}: Percent optimality gap for the Panwalkar-Smith-Koulamas (PSK) Heuristic \citep{PSK}
    \item \textbf{Pannerselvam}: Percent optimality gap for Pannerselvam's Heuristic \citep{Panneerselvam2006}
    \item \textbf{Optimal Value (Number of Instances Averaged)}: \(\checkmark\) Indicates the average of the optimal values for the total tardiness across instances. The term in parentheses, \textit{Number of Instances Averaged,} refers to the count of instances that are included in the average, specifically only those instances where a non-zero optimal value was obtained. The column \textit{Optimal Value} in Tables \ref{table:VIvsDP}--\ref{table:optimality_500_jobs} reports the optimal total tardiness for each instance class. These values correspond to the \textit{Optimal Tardiness Value} in Equation~\ref{eq:opt_gap}, serving as benchmarks for calculating optimality gaps and for assessing the relative performance of both heuristic and exact methods. Optimal values in Table \ref{table:VIvsDP} are computed directly using the DP formulation \eqref{eq:eq_12}--\eqref{eq:eq_15} for the 20-job instances. For 100, 200, and 500-job instances, optimal values are taken from \cite{shang2021}, where they are obtained using the Branch and Memorize algorithm.
\end{itemize}

The heuristics EDDC and MDDC were discovered using training instances of size $|N|=25$, and subsequently tested on larger problem sizes without retraining, in order to evaluate their scalability and generalization capacity.

Tables \ref{table:optimality20}--\ref{table:optimality_500_jobs} 
summarize algorithm performance for six representative classes and report overall 
averages across all 20 instance classes, comprising 800 problem instances in total minus instances with zero optimal value. According to the literature \cite{Potts1982}, the classes 
with RDD and TF values of $(0.2, 0.6)$ and $(0.2, 0.8)$ are the most challenging, 
while $(0.6, 0.4)$ and $(0.8, 0.8)$ correspond to medium difficulty, and 
$(0.2, 0.2)$ and $(0.4, 0.2)$ represent relatively easy instances. Each of the 
six rows in the tables corresponds to one of these representative classes. In 
addition, the ``Average'' row reports mean optimality gaps across all 20 parameter 
combinations (instance classes) with 40 instances in each class, covering the full set of 800 instances. Instances 
with zero optimal tardiness are excluded from these averages, as their inclusion 
would introduce division by zero in the optimality gap and distort the reported results. Such zero-optimal 
instances are correctly solved by all algorithms except Panneerselvam's Heuristic 
\citep{Panneerselvam2006}.

\begin{table}[h!]
    \caption{Optimality gaps of heuristic approaches for 20-job problems  where $n = |N|$ represents the number of jobs (\%)}
    \label{table:optimality20}
    \centering
    \resizebox{\textwidth}{!}{%
    \begin{threeparttable}
    \begin{tabular}{cccccccc}
    \hline
    \textbf{(n, RDD, TF)} & \textbf{EDD} & \textbf{EDDC} & \textbf{MDD} & \textbf{MDDC} & \textbf{Pannerselvam} & \textbf{PSK} & \textbf{Optimal Value} \\
    \hline
    (20, 0.2, 0.2) & 47.70 & 38.66 & 3.55 & 1.71 & 184.54 & 3.55 & 163 (40) \\
    (20, 0.2, 0.6) & 58.33 & 3.81 & 6.00 & 2.54 & 19.30 & 5.85 & 2255 (40) \\
    (20, 0.2, 0.8) & 62.56 & 1.57 & 2.93 & 0.47 & 7.23 & 2.90 &4179 (40) \\
    (20, 0.4, 0.2) & 4.98 & 4.76 & 0.70 & 4.66 &3218.31&0.69&37 (40)\\
    (20, 0.6, 0.4) &48.37&51.17&1.87&2.15&130.43&1.70&439 (40)\\
    (20, 0.8, 0.8) &52.89&1.87&0.28&0.26&19.71&0.28&4772 (40)\\
    \hline
    \textbf{Average} \footnote{}
    &46.61&12.83&1.86&1.70&411.49&1.77&1997 (680)\\
    \hline
    \end{tabular}
    \begin{tablenotes}
        \footnotesize
        \item \small 1 The six rows correspond to representative instance classes, while the ``Average'' row reports means across all 20 classes (800 instances, excluding zero-optimal cases). Optimal values are obtained using the Dynamic Programming approach for benchmarking.
    \end{tablenotes}
    \end{threeparttable}}
\end{table}

 Table \ref{table:optimality20} presents the performance comparison of various heuristics for 20-job problem instances with different characteristics. Each row in the table represents a problem class consisting of 40 instances generated using a uniform distribution as described in Section \ref{s:datagen}, with problem instances having zero optimal value excluded. With six representative classes shown, the table summarizes the performance of the algorithms across a total of 800 instances, highlighting their effectiveness at varying levels of difficulty. Classes with RDD and TF values of (0.2, 0.6) and (0.2, 0.8) represent the most challenging instances, as identified in the literature (\cite{Potts1982}). Meanwhile, the (0.6, 0.4) and (0.8, 0.8) classes are of medium difficulty, and (0.2, 0.2) and (0.4, 0.2) represent relatively easy instance classes. The ``Average'' row reports the average optimality gaps across 20 parameter combinations (instance classes), totaling 800 problem instances, excluding the zero optimal value instances which are solved correctly by all of the algorithms except Pannerselvam.

As shown in  Table \ref{table:optimality20}, the EDD Challenger (EDDC) outperforms both the original EDD algorithm and Pannerselvam’s heuristic for the 20-job problems. However, the MDD Challenger (MDDC) consistently achieves lower optimality gaps than EDDC across all demonstrated instance classes and in the overall results. Although EDDC performs better than Pannerselvam's heuristic and EDD, it does not surpass the performance of PSK, MDD, and MDDC, which closely follow each other. MDDC achieves the best performance across all problem instances compared to existing algorithms in this problem size. It particularly excels in solving harder problem instances, although it fails to deliver robust performance in some instance classes such as (0.4, 0.2). Thus, MDDC’s advancement over the state-of-the-art is primarily driven by its superior performance on the hardest instances. The highest optimality gaps are observed for the (0.2, 0.6), (0.2, 0.8), and (0.2, 0.2) classes. The lower performances on the (0.2, 0.2) and (0.4, 0.2) classes are attributed to their low tardiness values, where the mis-assignment of just one or two jobs can cause significant shifts in the optimality gap percentage. In summary, MDDC is the best performing method for this problem scale, closely followed by PSK and then MDD. EDDC ranks next, surpassing Pannerselvam’s heuristic despite its limitations, demonstrating its value.

\begin{table}[h!]
    \caption{Optimality gaps of heuristic approaches for 20-job problems with out-of-sample dataset where $n = |N|$ represents the number of jobs (\%)}
    \label{table:optimality_outofsample}
    \centering
    \resizebox{\textwidth}{!}{%
    \begin{threeparttable}
    \begin{tabular}{cccccccc}
    \hline
    \textbf{(n, RDD, TF)} & \textbf{EDD} & \textbf{EDDC} & \textbf{MDD} & \textbf{MDDC} & \textbf{Pannerselvam} & \textbf{PSK} & \textbf{Optimal Value} \\
    \hline
    (20, 0.2, 0.2) & 29.50 & 37.90 & 5.57 & 1.25 & 151.13 & 4.34 & 198 (40) \\
    (20, 0.2, 0.6) & 36.68 & 4.20 & 5.94 & 2.87 & 22.70 & 5.82 & 3016 (40) \\
    (20, 0.2, 0.8) & 37.15 & 1.51 & 2.32 & 0.72 & 11.73 & 2.32 & 5751 (40) \\
    (20, 0.4, 0.2) & 1.38 & 1.38 & 0.72 & 0.60 & 3506.00 & 0.72 & 28 (40) \\
    (20, 0.6, 0.4) & 28.70 & 37.15 & 3.84 & 2.91 & 154.41 & 3.84 & 479 (40) \\
    (20, 0.8, 0.8) & 35.76 & 1.40 & 0.25 & 0.23 & 25.16 & 0.25 &5725 (40) \\
    \hline
    \textbf{Average} \footnote{}
    &31.94&11.30&2.51&1.71&366.68&2.41&2645 (670)\\
    \hline
    \end{tabular}
    \begin{tablenotes}
        \footnotesize
        \item \small 2 The six rows correspond to representative instance classes, while the ``Average'' row reports means across all 20 classes (800 instances, excluding 130 zero-optimal cases). Optimal values are obtained using the Dynamic Programming approach for benchmarking.
    \end{tablenotes}
    \end{threeparttable}}
\end{table}

   Table \ref{table:optimality_outofsample} presents the performance of the algorithms on an out-of-sample dataset generated using a normal distribution (see Section \ref{s:datagen}). The ranking of the algorithms remains consistent with the first dataset for this job size, though slight variations in optimality gaps are observed. While MDD, MDDC and PSK continue to perform best, their optimality gaps increase. EDD and Pannerselvam's algorithm are again surpassed by EDDC, having a lower optimality gap compared to the first dataset. These results demonstrate the robust performance of MDDC compared to other state-of-the-art algorithms.

\begin{table}[h!]
    \caption{Optimality gaps of heuristic approaches for 100-job problems where $n = |N|$ represents the number of jobs (\%)}
    \label{table:optimality_100}
    \centering
    \resizebox{\textwidth}{!}{%
    \begin{threeparttable}
    \begin{tabular}{cccccccc}
    \hline
    \textbf{(n, RDD, TF)} & \textbf{EDD} & \textbf{EDDC} & \textbf{MDD} & \textbf{MDDC} & \textbf{Pannerselvam} & \textbf{PSK} & \textbf{Optimal Value} \\
    \hline
    (100, 0.2, 0.2) & 50.87 & 73.22 & 3.87 & 3.05 & 343.45 & 3.87 & 2291 (10) \\
    (100, 0.2, 0.6) & 67.76 & 6.21 & 5.44 & 5.23 & 27.84 & 5.37 & 49367 (10) \\
    (100, 0.2, 0.8) & 63.64 & 1.88 & 3.02 & 2.82 & 10.04 & 2.98 & 97196 (10) \\
    (100, 0.4, 0.2) & 7.59 & 7.59 & 0.00 & 0.00 & 33541.59 & 0.00 & 34 (9) \\
    (100, 0.6, 0.4) & 66.68 & 108.53 & 2.77 & 1.65 & 308.33 & 2.77 & 5267 (10) \\
    (100, 0.8, 0.8) & 62.49 & 0.50 & 0.02 & 0.02 & 27.37 & 0.02 &93103 (10) \\
    \hline
    \textbf{Average} \footnote{} 
    &59.57&28.12&1.81&1.52&3433.60&1.79&41909 (163)\\
    \hline
    \end{tabular}
    \begin{tablenotes}
        \footnotesize
        \item \small 3 The six rows correspond to representative instance classes, while the ``Average'' row reports means across all 20 classes (200 instances, excluding 37 zero-optimal cases). Optimal values are obtained from \cite{shang2021} for benchmarking.
    \end{tablenotes}
    \end{threeparttable}}
\end{table}

Table \ref{table:optimality_100} shows that PSK and MDD perform better in 100-job instances, exhibiting lower optimality gaps than in 20-job instances, especially in harder problems. However, MDDC performs better than the other algorithms, having the lowest optimality gap, although the optimality gap increases for hard instances. The optimality gaps of EDD, EDDC, and Pannerselvam also increase, but their relative rankings in the performance hierarchy remain unchanged. It should be noted that Pannerselvam's algorithm demonstrates poorer performance compared to the first two datasets, losing its value with increasing problem size.

\begin{table}[h!]
    \caption{Optimality gaps of heuristic approaches for 200-job problems  where $n = |N|$ represents the number of jobs (\%)}
    \label{table:optimality_200_jobs}
    \centering
    \resizebox{\textwidth}{!}{%
    \begin{threeparttable}
    \begin{tabular}{ccccccc}
    \hline
    \textbf{(n, RDD, TF)} & \textbf{EDD} & \textbf{EDDC} & \textbf{MDD} & \textbf{MDDC} & \textbf{PSK} & \textbf{Optimal Value} \\
    \hline
    (200, 0.2, 0.2) & 66.09 & 83.70 & 6.04 & 5.83 & 6.04 & 8429 (10) \\
    (200, 0.2, 0.6) & 68.16 & 5.57 & 5.40 & 5.33 & 5.40 & 194810 (10) \\
    (200, 0.2, 0.8) & 63.15 & 1.72 & 2.83 & 2.73 & 2.83 & 383119 (10) \\
    (200, 0.4, 0.2) & 11.17 & 11.17 & 0.00 & 0.00 & 0.00 & 23 (9) \\
    (200, 0.6, 0.4) & 58.17 & 125.88 & 1.81 & 1.68 & 1.80 & 16152 (10) \\
    (200, 0.8, 0.8) & 68.32 & 0.36 & 0.02 & 0.02 & 0.02 &330949 (10) \\
    \hline
    \textbf{Average} \footnote{}
    &62.98&52.02&2.07&2.00&1.99&164883 (160)\\
    \hline
    \end{tabular}
    \begin{tablenotes}
        \footnotesize
        \item \small 4 The six rows correspond to representative instance classes, while the ``Average'' row reports means across all 20 classes (200 instances, excluding 40 zero-optimal cases). Optimal values are obtained from \cite{shang2021} for benchmarking.
    \end{tablenotes}
    \end{threeparttable}}
\end{table}

As shown in Tables \ref{table:optimality_200_jobs} and \ref{table:optimality_500_jobs}, MDDC and competing state-of-the-art algorithms exhibit similar optimality gaps for larger problem sizes (200 and 500 jobs). Here, we omit Pannerselvam's results since they have already found to be the least effective for instances with smaller jobs. While PSK, MDD, and MDDC perform worse on harder problems, PSK and MDD compensate with better performance on medium-hardness instances. In contrast, MDDC improves its performance for (0.2, 0.8) problems, demonstrating adaptability. The convergence of optimality gaps among MDDC, MDD, and PSK indicates their scalability and robust performance across larger problem sizes. For 500-job problems, PSK, MDD, and MDDC exhibit higher optimality gaps due to the increased problem size. Nevertheless, MDDC remains the most effective among the algorithms.

Overall, the EDD, EDDC, and Pannerselvam heuristics demonstrate limited scalability, with increasing optimality gaps as the problem size grows. Although EDD suffers from the increased problem scale, it outperforms EDDC in 500-job problems. Among these algorithms, Pannerselvam is the one with the highest optimality gap, completely losing its appeal for implementation; hence, it is not included in the comparison. MDDC, in particular, consistently outperforms EDDC on all scales of the problem, reflecting the impact of different initial solutions provided to LLM during the start of the heuristic discovery pipeline.

Moreover, EDDC and MDDC exhibit improved optimality gaps compared to their predecessors, EDD and MDD, albeit with increased complexity. This complexity results in greater implementation challenges and longer run times, presenting a trade-off between accuracy and complexity when selecting an algorithm for implementation.

\begin{table}[h!]
    \caption{Optimality gaps of heuristic approaches for 500-job problems  where $n = |N|$ represents the number of jobs (\%)}
    \label{table:optimality_500_jobs}
    \centering
    \resizebox{\textwidth}{!}{%
    \begin{threeparttable}
    \begin{tabular}{ccccccc}
    \hline
    \textbf{(n, RDD, TF)} & \textbf{EDD} & \textbf{EDDC} & \textbf{MDD} & \textbf{MDDC} & \textbf{PSK} & \textbf{Optimal Value} \\
    \hline
    (500, 0.2, 0.2) & 69.29 & 94.04 & 7.91 & 7.26 & 7.91 & 48316 (10) \\
    (500, 0.2, 0.6) & 67.12 & 6.91 & 5.47 & 5.38 & 5.46 & 1216062 (10) \\
    (500, 0.2, 0.8) & 65.38 & 2.11 & 3.36 & 3.28 & 3.36 & 2350056 (10) \\
    (500, 0.4, 0.2) &120.20&120.20&98.99&98.99&98.99&38 (10)\\
    (500, 0.6, 0.4) &81.12&144.23&3.29&3.23&3.29&92023 (10)\\
    (500, 0.8, 0.8) &70.10&0.36&0.25&0.24&0.25&2054039 (10)\\
    \hline
    \textbf{Average} \footnote{}
    &127.50&271.75&48.75&48.40&48.74&998667 (160)\\
    \hline
    \end{tabular}
    \begin{tablenotes}
        \footnotesize
        \item  \small 5 The six rows correspond to representative instance classes, while the ``Average'' row reports means across all 20 classes (200 instances, excluding 40 zero-optimal cases). Optimal values are obtained from \cite{shang2021} for benchmarking.
    \end{tablenotes}
    \end{threeparttable}}
\end{table}

As Table \ref{table:solution_times} displays, EDDC and MDDC provide significant computational efficiency compared to the Branch \& Memorize algorithm proposed by \cite{shang2021}. Across 200 problem instances for both 200-job and 500-job problems which were provided by \cite{shang2021}, EDDC and MDDC consistently achieve significantly shorter average solution times. For example, MDDC solves 200-job problems in 0.00013 CPU seconds on average and 500-job problems in approximately 0.00081 CPU seconds on average, compared to 0.02039 and 1.12891 CPU seconds on average by the Branch \& Memorize of \cite{shang2021} algorithm implemented in our setup, respectively. EDDC's solution times are close to MDDC with higher solution times. Although the optimality gaps are higher, this demonstrates the computational advantage of the newly discovered algorithms.

\begin{table}[h!]
    \caption{Comparison of solution times (average\ CPU seconds) between MDDC, EDDC, and Branch \& Memorize~\cite{shang2021}}
    \label{table:solution_times}
    \centering
    \resizebox{\textwidth}{!}{%
    \begin{threeparttable}
    \begin{tabular}{ccccccc}
    \hline
    \textbf{Number of jobs (n)} & \textbf{MDDC (s)} & \textbf{EDDC (s)} & \textbf{Branch \& Memorize (s)} \\
    \hline
    200 & 0.00013 & 0.00015 & 0.02039 \\
    500 & 0.00081 & 0.00190 & 1.12891 \\
    \hline
    
    \hline
    \end{tabular}
    \end{threeparttable}}
\end{table}

This efficiency makes MDDC particularly preferable for smaller problem sizes, achieving lower optimality gaps of approximately 2\% while providing significantly faster computation compared to exact methods. As problem sizes grow, the challenges become even more pronounced; for example, \( N \geq 100 \) instances are unsolvable using the state-of-the-art MIP solver, Gurobi V11.0.1's Branch \& Cut algorithm. In such cases, the simplicity and computational efficiency of the LLM-driven heuristics EDDC and MDDC offer clear advantages over more complex exact algorithms and solvers, including Branch \& Memorize. These advantages are particularly evident in terms of computational efficiency and ease of implementation, making the heuristics more practical for large and complex problem instances. Simple heuristics typically require less computational resources and are faster to execute, which is crucial in practical settings where quick decision making is essential. This practicality is especially beneficial in operational environments where the slightly higher optimality gaps of the heuristics are acceptable trade-offs for significant gains in computational speed and reduced implementation complexity.


In summary, MDDC consistently outperforms other algorithms across all problem sizes except for the 200-job instances, underscoring the effectiveness of the pipeline. Its closest competitors are PSK, followed by MDD. Although all algorithms exhibit increasing optimality gaps as problem size grows, EDDC generally performs better than its baseline, EDD, except in the 500-job cases, where EDDC's optimality gap is approximately twice as large as that of EDD. These findings suggest that while the learned heuristics scale reasonably well to medium-sized problems, their performance on larger instances may be constrained by limited training data from smaller problem sizes. Nevertheless, two of the generated heuristics remain competitive, with MDDC standing out as the top-performing algorithm overall, aside from a narrow gap in the 200-job problem set.

\subsubsection{Performance Analysis of Augmented MDDC}

Algorithm~\ref{alg:AugmentedMDDC} presents the Augmented MDDC algorithm which is developed based on \cite{Bean}'s Augmented MDD heuristic. This algorithm incorporates a local search step at each iteration, immediately after scheduling a job with the MDDC heuristic. By introducing local improvements throughout the scheduling process, the algorithm gains a more global perspective, which helps reduce the optimality gap. Table~\ref{table:optimality_scale} compares the performance of MDD, MDDC, the original Augmented MDD algorithm (\cite{Bean}), and the augmented MDDC.

\begin{algorithm}
    \caption{Augmented MDDC Algorithm: Iteratively schedules jobs using the MDDC
    rule, followed by local search refinements.}
    \label{alg:AugmentedMDDC}
    \begin{algorithmic}[1]
        \State \textbf{Input:} Processing times $p$ and due dates $d$ for each job
        \State \textbf{Output:} Schedule $S$
        \Procedure{AugmentedMDDC}{$p$, $d$}
            \State Initialize $current\_time \gets 0$, $S \gets [\,]$ (empty sequence)
            \State Let $U \gets$ set of all unscheduled jobs
            \State Sort $U$ based on Shortest Processing Time (SPT) rule
            \State Initialize $p_U$ and $d_U$ as the processing times and due dates of all unscheduled jobs

            \While{$U \neq \emptyset$}
                \State $\mu \gets \max\left(p_U \, \mathbf{\times \, \textbf{1.1}} + current\_time, d_U\right)$
                \State $\boldsymbol{\rho \gets \min\left(\frac{p_U}{current\_time + \max(p_U)}, 1\right)}$
                \State $\boldsymbol{\theta \gets \frac{\rho^2}{1 + \rho^2}}$
                \State $\boldsymbol{\mu \gets \mu \times (1 + \theta)}$
                \State $\boldsymbol{\sigma \gets \frac{p_U}{current\_time + \sum(p_U) / |U|}}$
                \State $\boldsymbol{\mu \gets \mu + \sigma}$

                \State \textbf{Select job $j^*$ with the minimum $\boldsymbol{\mu}$ score and append it to $S$}
                \State \textbf{Remove $j^*$ from $U$}
                \State \textbf{Update $current\_time \gets current\_time + p_{j^*}$}
                \State \textbf{Update $p_U$, $d_U$ to reflect removal of $j^*$}

                \For{each job $j \in S \setminus \{j^*\}$}
                    \State Compute current total tardiness; set $best\_tardiness \gets$ value
                    \State Construct a trial schedule $S'$ by moving job $j$ to the end of $S$
                    \State Compute tardiness of $S'$ as $tardiness$
                    \If{$tardiness < best\_tardiness$}
                        \State $S \gets S'$
                        \State $best\_tardiness \gets tardiness$
                    \EndIf
                \EndFor
            \EndWhile
            \State \Return $S$
        \EndProcedure
    \end{algorithmic}
\end{algorithm}

These algorithms are evaluated separately, as the augmented versions incorporate local search, improving accuracy at the cost of increased computational complexity and longer solution times. Although the augmented versions outperform their standard counterparts, the margin of improvement diminishes as the size of the problem increases. In particular, MDDC achieves lower optimality gaps than augmented MDD, likely due to the targeted training process of MDDC on smaller instances. Our final observation is that augmented MDDC consistently achieves the best performance across all problem sizes considered in this study.

\begin{table}[h!]
    \caption{Optimality gaps of heuristics across different job sizes (\%)}
    \label{table:optimality_scale}
    \centering
    \resizebox{\textwidth}{!}{%
    \begin{tabular}{ccccc}
    \hline
    Number of Jobs (n) & MDD & Augmented MDD & MDDC & Augmented MDDC \\
    \hline
    20 & 1.86 & 1.21 & 1.70 & 0.79 \\
    100 & 1.81 & 1.27 & 1.52 & 1.08 \\
    200 & 2.07 & 1.53 & 2.00 & 1.44 \\
    500 & 48.75 & 48.15 & 48.40 & 47.70 \\  
    \hline
    \end{tabular}}
\end{table}

Table \ref{table:solution_times_scale} presents a comparison of the solution times for MDD, MDDC, and their augmented versions. The augmented algorithms demand more computation, since after each scheduling step \(n\), the algorithm reevaluates \(n-1\) remaining jobs for potential reassignments. MDDC consistently requires more time than MDD \textcolor{blue}{with close differences}, both in its standard and augmented forms, which we attribute to the more intricate sequence of arithmetic operations involved in MDDC. Although the difference in solution times between MDD and MDDC is \textcolor{blue}{considerable}, improvements in optimality gaps, especially for smaller instances, are significant and proves that LLM-inspired modifications are worthwhile additions to the algorithm. This trade-off underscores the value of MDDC in scenarios where solution accuracy is prioritized over computational efficiency.

\begin{table}[h!]
    \caption{Average solution times across different job sizes (CPU seconds)}
    \label{table:solution_times_scale}
    \centering
    \resizebox{\textwidth}{!}{%
    \begin{tabular}{ccccc}
    \hline
    Number of Jobs (n) & MDD & Augmented MDD & MDDC & Augmented MDDC \\
    \hline
    100 & 0.00001 & 0.00039 & 0.00004 & 0.00041 \\
    200 & 0.00003 & 0.00258 & 0.00013 & 0.00265 \\
    500 & 0.00017 & 0.03626 & 0.00081 & 0.03638 \\  
    \hline
    \end{tabular}}
\end{table}

\section{Conclusion}\label{s:conclusion}
In this study, we leverage the power and flexibility of LLMs through island-based evolution for discovering heuristics to solve the single-machine scheduling problem, aiming to minimize total tardiness. As a result, we introduce two novel heuristics: EDD Challenger (EDDC) and MDD Challenger (MDDC). These heuristics build on established algorithms—EDD and MDD—and represent significant innovations in heuristic development.  We compare the performance of LLM-driven heuristics with a range of exact and heuristic methods. MDDC advances the state-of-the-art by consistently outperforming traditional and other LLM-driven heuristics, demonstrating superior scalability and adaptability across problem sizes. Moreover, EDDC, enhanced with LLM improvements, outperforms both the original EDD and Pannerselvam's algorithms for job sizes up to 500. However, it falls short of the leading heuristics in accuracy, such as MDD and MDDC, highlighting its limitations. Both EDDC and MDDC outperform Gurobi in terms of computation time across every problem size, as these exact methods are not computationally tractable for large problem sizes.
 
Although trained on datasets comprising 25-job problems, the developed heuristics, particularly MDDC, generalize effectively to larger problem instances without any modifications, further enhancing their practical value. Furthermore, we prove that by applying an efficient, single-threaded approach (compared to \cite{funsearch}'s resource intensive methodology), it is possible to discover heuristics if we ensure an effective implementation backed up by problem-specific specification and evaluation, prompt engineering, and hallucination control. This study exemplifies the synergy between the brute-force capability of LLMs and the understanding and problem-solving capabilities of human experts. By translating the problem structure into computer code, we create a sandbox environment where LLM iteratively refines solutions through cumulative improvements. Although the results are not interpretable, they perform well against the heuristics based on the literature. This points out a further direction: using explainable and interpretable algorithms such as tree-based algorithms alongside LLMs.

Our findings verify that the iterative heuristic development method presented in this study produces impactful results. Although the heuristic discovery process is computationally intensive - requiring up to 72 CPU hours - it represents a one-time investment that produces reusable algorithms capable of delivering superior performance in a wide range of hard scheduling problems. These algorithms not only contribute new insights to the literature, but can also be immediately deployed to solve numerous practical instances with negligible additional cost. In particular, the best-performing algorithms in both experiments often emerge well before the training process is complete, suggesting that a portion of the remaining iterations may be redundant. As a future direction, this presents an opportunity to enhance training efficiency by incorporating adaptive sampling strategies or more targeted stopping criteria to avoid unnecessary computation. This research serves as a pivotal demonstration of the efficacy of collaboration between LLM and human expertise. And moreover, this work not only showcases the ability of LLMs to contribute to existing algorithms through the introduction of regression-similar terms, but also highlights their role as sources of inspiration, complementing human expertise.

Looking ahead, this methodology opens exciting avenues for future research with the potential to address a broader range of mathematical problems, particularly within combinatorial optimization and mixed-integer programming. Especially, extending our approach to various combinatorial problems such as job-shop scheduling, capacitated lot-sizing, and vehicle routing promises potential, given that the algorithms generated with this methodology managed to advance the state of the art. By continuing to explore LLM-based methods, our aim is to inspire new heuristic and algorithm developments and further advance the frontier of algorithmic optimization and the solution of NP-Hard problems.

\section*{Code and Data Availability}

All code and data supporting the results of this study are available at:\\
\url{https://github.com/ibrahimoguzc/DiscoverHeuristics}

\section*{Acknowledgments} 
We gratefully acknowledge the support of the National Science Foundation CAREER Award co-funded by the CBET/ENG Environmental Sustainability program and the Division of Mathematical Sciences in MPS/NSF under Grant No. CBET-1554018. The authors are grateful for the funding provided by the Grado Department of Industrial and Systems Engineering and the computing clusters offered by the Advanced Computing Resources (ARC) Center at Virginia Tech. We sincerely thank the four reviewers, the associate editor, and the editor for their insightful comments and suggestions, which have helped us substantially improve the clarity, depth, and overall quality of the manuscript.

\Needspace{12\baselineskip}   

\section*{Appendix}
\Needspace{0.6\textheight} 
\setcounter{figure}{0}
\renewcommand{\thefigure}{A\arabic{figure}}

\begin{figure}[H]
  \centering
  \includegraphics[width=1\linewidth,height=0.95\textheight,keepaspectratio]{\detokenize{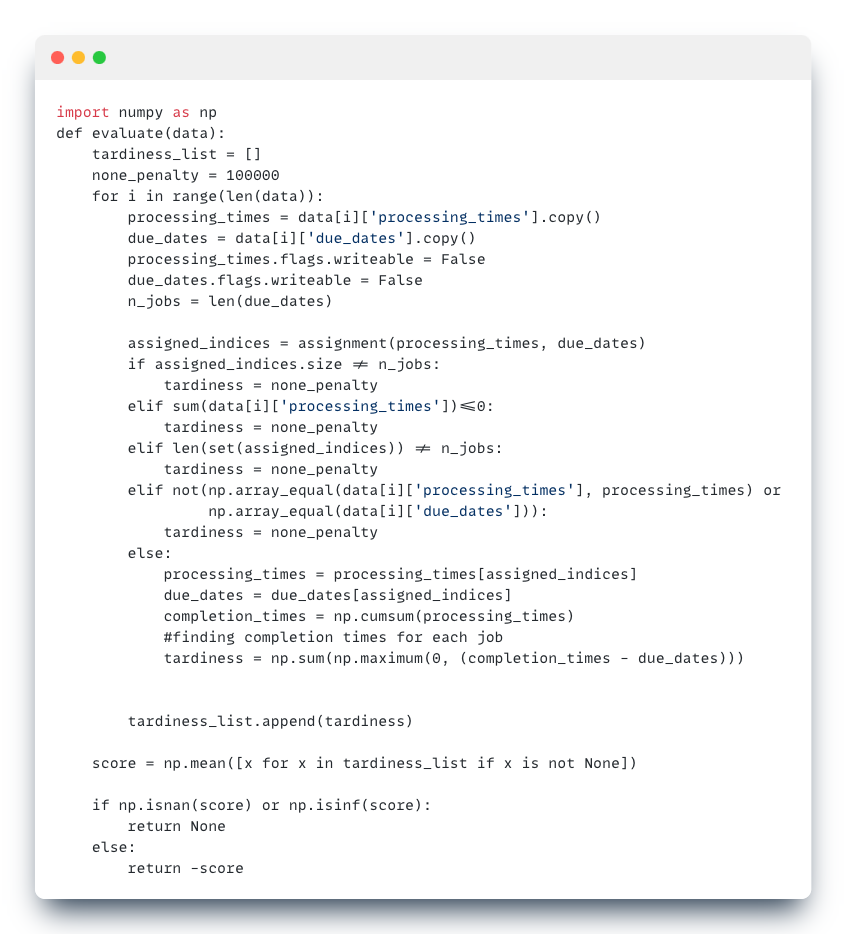}}
  \caption{`Evaluate' function for SMTT problem}
  \label{fig:evaluate_figure}
\end{figure}

\begin{figure}
\centering
\includegraphics[width=0.8\linewidth]{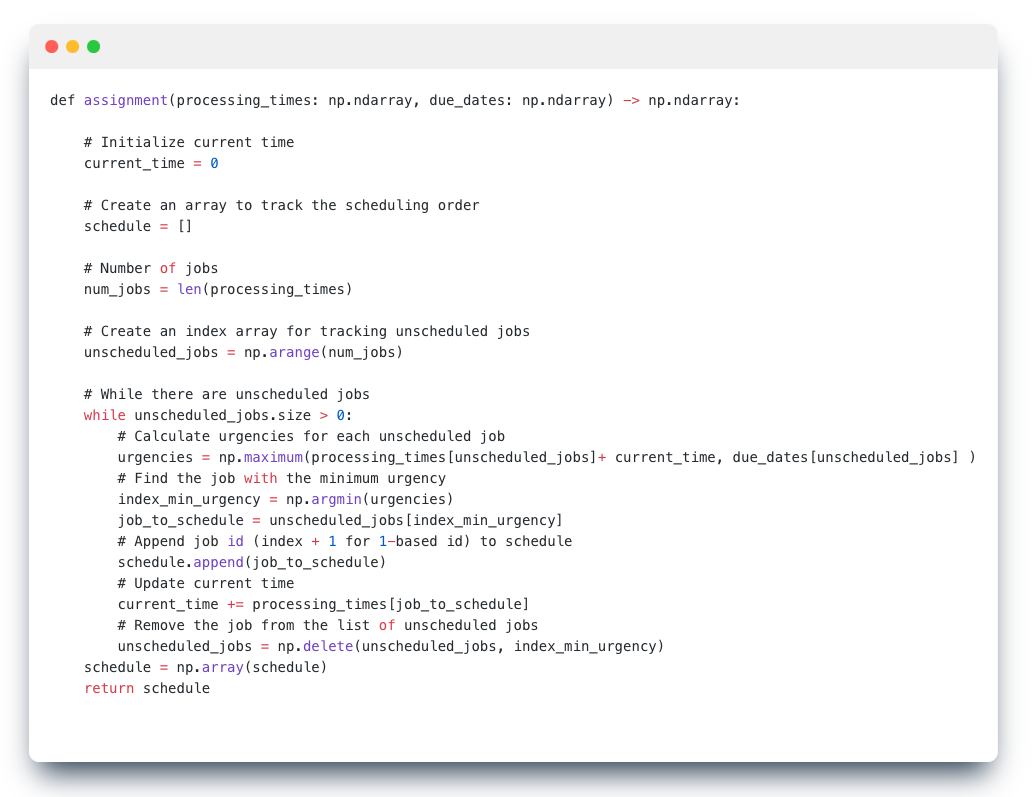}
\caption{`Assignment' function for SMTT problem}
\label{fig:assignment_figure}
\end{figure}

\begin{figure}
\centering
\includegraphics[width=0.9\linewidth]{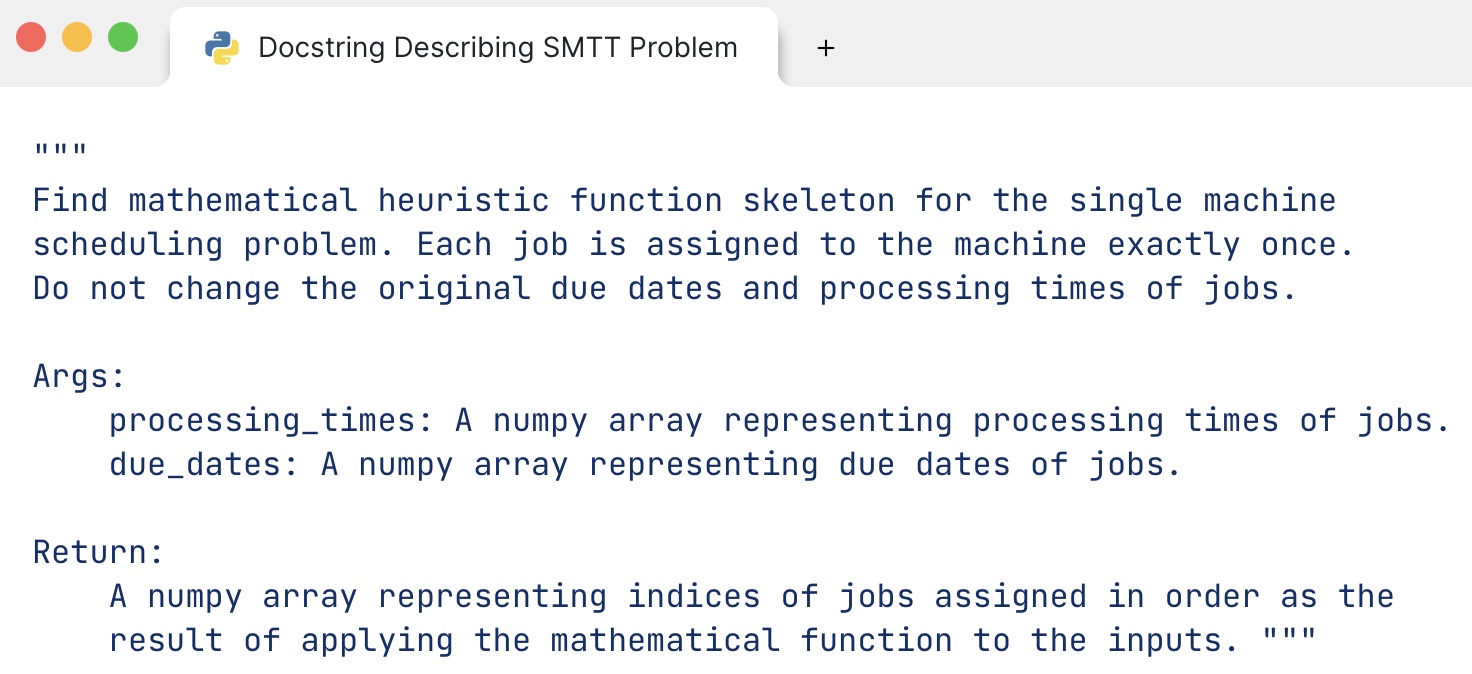}
\caption{Docstring provided to LLM for SMTT problem}
\label{fig:docstring_figure}
\end{figure}

\if0\blind{

\newpage
\bibliographystyle{chicago}
\spacingset{1}
\bibliography{IISE-Trans}
	
\end{document}